%% file: acl_2025.tex
\pdfoutput=1

\documentclass[11pt]{article}

\usepackage[preprint]{acl_2025}
\usepackage{times}
\usepackage{latexsym}

\usepackage[T1]{fontenc}

\usepackage[utf8]{inputenc}

\usepackage{microtype}

\usepackage{inconsolata}

\usepackage{graphicx}

\usepackage{hyperref}
\usepackage{url}
\usepackage[utf8]{inputenc} %
\usepackage[T1]{fontenc}    %
\usepackage{longtable}
\usepackage{booktabs}       %
\usepackage{amsfonts}       %
\usepackage{nicefrac}       %
\usepackage{colortbl}
\usepackage{tocloft}
\usepackage{titletoc}
\usepackage{titlesec}
\usepackage{soul}
\usepackage{paralist}       %
\usepackage{graphicx}
\usepackage{amsmath}
\usepackage[ruled,vlined]{algorithm2e}
\usepackage{caption}
\usepackage{subcaption}
\usepackage{diagbox}
\usepackage{bbding}
\usepackage{multirow}
\usepackage{pifont}
\usepackage{wrapfig}
\usepackage{lipsum}
\usepackage{capt-of}    %
\usepackage{tabularx}   %
\usepackage{multirow}
\usepackage{makecell}
\usepackage{bbm}
\usepackage{pdfpages}
\usepackage{array, makecell} %

\usepackage{float}
\usepackage{authblk}
\usepackage{placeins}
\usepackage{enumitem}

\newcommand{\react}{\texttt{ReAct}}

\newcommand{\respact}{\texttt{ReSpAct}}
\definecolor{sectioncolor}{RGB}{0,90,160}
\definecolor{subsectioncolor}{RGB}{100,100,100}

\title{ReSpAct: Harmonizing Reasoning, Speaking, and Acting \\ Towards Building Large Language Model-Based Conversational AI Agents}

\author{\textbf{Vardhan Dongre, Xiaocheng Yang, Emre Can Acikgoz,} \\
\textbf{Suvodip Dey, Gokhan Tur, Dilek Hakkani-Tür} \\
  University of Illinois Urbana-Champaign \\
  \texttt{\{vdongre2,xy61,acikgoz2,sdey,gokhan,dilek\}@illinois.edu} \\
}

\begin{document}
\maketitle

\input{acl_2025_sections/text/abstract}

\begin{figure}[t]
\centering
\includegraphics[width=\linewidth]{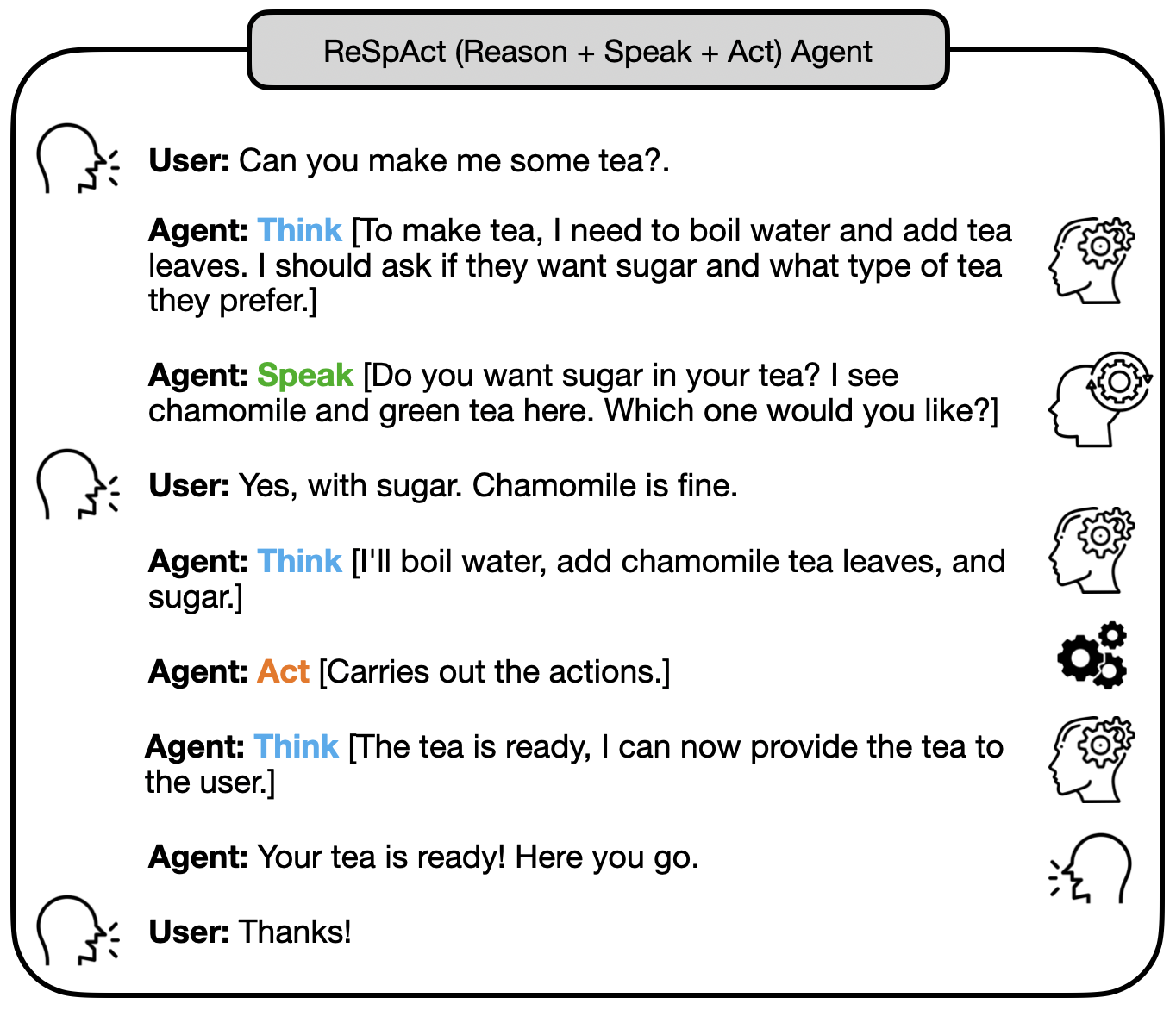}
\vspace*{-4ex}
\caption{\textbf{\respact{}} is a framework for task-oriented conversational agents that allows agents to ask questions, request feedback, and adapt their strategies based on user input. }
\label{fig:respact_agent}
\vspace*{-4ex}
\end{figure}

\begin{figure*}[t]
\centering
\includegraphics[width=\textwidth]{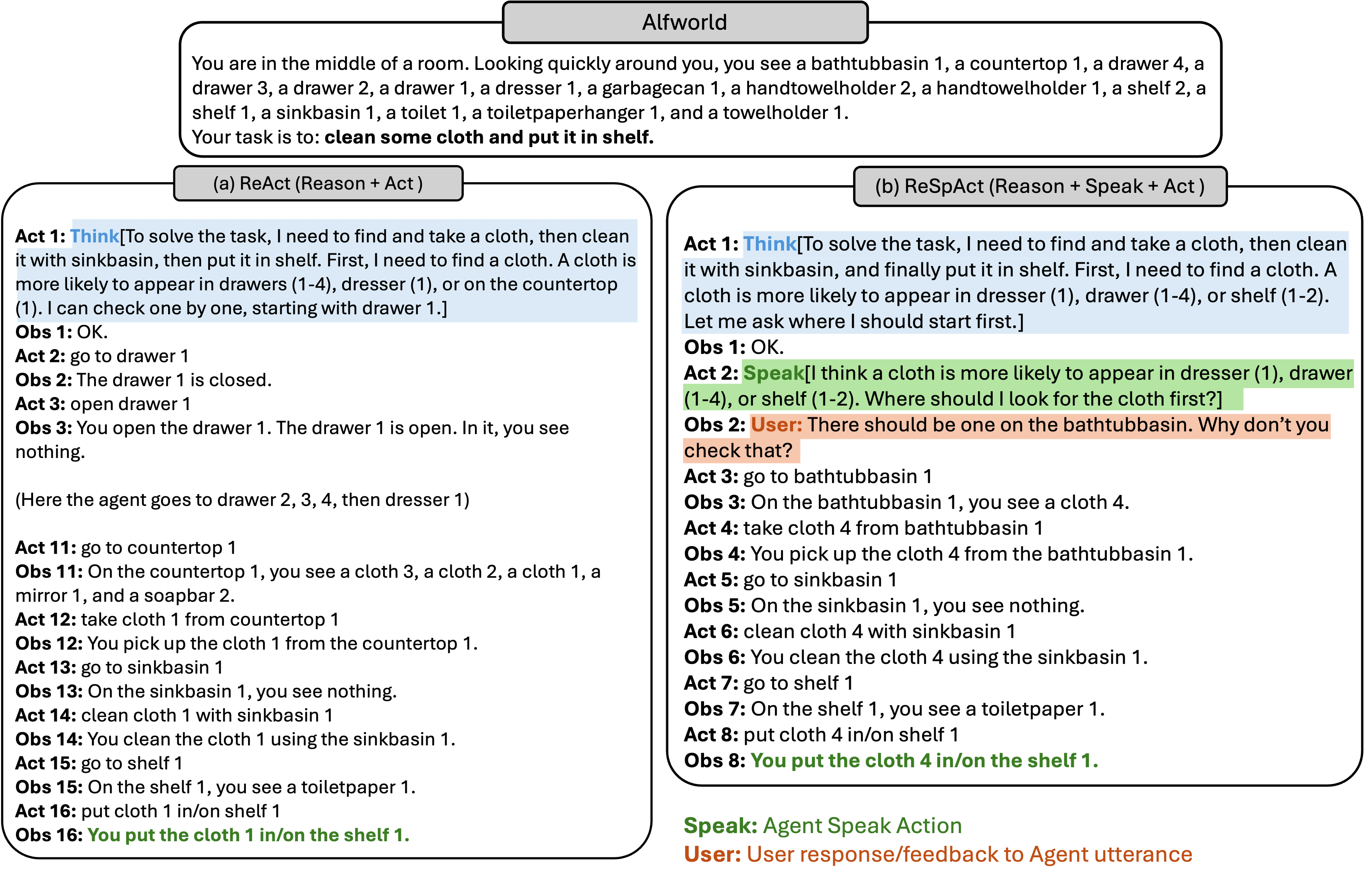}
\vspace*{-4ex}
\caption{Comparison of (a) ReAct and (b) ReSpAct to solve a game in AlfWorld \cite{shridhar2020alfworld}. We show only the task-solving trajectories generated by the model (Act, Thought and Speech) and the environment (Obs). 
}
\label{fig:overview}
\vspace*{-2ex}
\end{figure*}

\vspace*{-2ex}
\input{acl_2025_sections/text/introduction}
\vspace*{-2ex}
\input{acl_2025_sections/text/related_work}
\vspace*{-1ex}
\input{acl_2025_sections/text/methods}

\vspace*{-0ex}
\input{acl_2025_sections/text/experiments}
\input{acl_2025_sections/text/results}
\input{acl_2025_sections/text/discussions}
\vspace*{-1ex}
\input{acl_2025_sections/text/conclusions}
\input{acl_2025_sections/text/limitations}

\section{Impact Statement}
The \respact{} framework improves LLM-based agents by enabling interactive, policy-guided action determination while keeping humans in the loop. This approach enhances collaboration and task success by ensuring agents seek clarification and guidance rather than acting on assumptions. However, increasing agent autonomy may introduce risks, such as over-reliance or security concerns in sensitive environments. \respact{} mitigates these risks by emphasizing human involvement and dynamic dialogue, promoting better alignment and safety. Further research is needed to explore potential challenges and ensure responsible AI use.

\section{Acknowledgments}

This work was supported in part by Other Transaction award HR0011249XXX from the U.S. Defense Advanced Research Projects Agency (DARPA) Friction for Accountability in Conversational Transactions (FACT) program and has benefited from the Microsoft Accelerate Foundation Models Research (AFMR) grant program, through which leading foundation models hosted by Microsoft Azure and access to Azure credits were provided to conduct the research.

\bibliography{acl_2025}
\input{acl_2025_sections/text/appendix}

\end{document}

%% file: acl_2025_sections/text/abstract.tex
\begin{abstract}

Large language model (LLM)-based agents are increasingly employed to interact with external environments (e.g., games, APIs, world models) to solve user-provided tasks. However, current frameworks often lack the ability to collaborate effectively with users in fully conversational settings. Conversations are essential for aligning on task details, achieving user-defined goals, and satisfying preferences. While existing agents address ambiguity through clarification questions \cite{li2023eliciting, zhang2023clarify, chen2023asking}, they underutilize the broader potential of a LLM's conversational capabilities. In this work, we introduce \respact{}, an LLM-based agent designed to seamlessly integrate reasoning, decision-making, and dynamic dialogue for task-solving. Expanding on reasoning-first approaches like \react{} \cite{yao2022react}, \respact{} employs active, free-flowing dialogues to interpret instructions, clarify goals, provide status updates, resolve sub-task failures, and refine plans based on user inputs without any explicit dialogue schema. By alternating between task-solving actions and interactive conversations, \respact{} demonstrates improved performance across diverse environments. We evaluate \respact{} in user-interactive settings, including task-oriented dialogue systems (MultiWOZ) and decision-making tasks (Alfworld, WebShop). \respact{} outperforms \react{} with absolute success rate improvements of 6\% and 4\% in Alfworld and WebShop, respectively, and achieves a 5.5\% gain in Inform and a 3\% gain in Success scores in MultiWOZ. These results highlight the value of integrating dynamic user-agent collaboration for more effective task resolution.

\end{abstract}

%% file: acl_2025_sections/text/introduction.tex
\section{Introduction}
Instruction-following is a fundamental capability for intelligent agents operating in real-world environments. Recent works such as \cite{wei2022chain, huang2022inner, yao2022react, shinn2024reflexion} have focused primarily on building agents that can follow individual instructions without considering the importance of feedback and interaction. In realistic settings, instruction-following often involves a back-and-forth exchange between the agent and the user to reduce uncertainties, correct mistakes, and handle exceptions \cite{dai2024think}.

Effective conversational agents go beyond clarifying ambiguities—they actively collaborate with users by offering alternative suggestions, providing status updates, and following up on requests to ensure alignment with user goals. For example, when asked to “Go to the kitchen and bring me the pan,” an agent can confirm which pan is needed if multiple options exist, suggest alternatives if the desired pan is unavailable, and update the user on progress. Similarly, when tasked with “Arrange a trip to Hawaii,” the agent can verify key details, propose travel options based on preferences, and keep the user informed throughout the process. This dynamic interaction enables agents to adapt to evolving user needs, ensuring tasks are completed efficiently and effectively. 


Existing reasoning and decision-making approaches for language agents augment the agent's action space with a language model, allowing the agent to generate free-form thoughts in natural language that help contextualize and reason about the task at hand. By alternating between task-solving actions and language thoughts, these agents can perform multi-step reasoning and compose useful information for solving complex tasks. However, such frameworks do not explicitly incorporate user interaction and feedback into the agent's reasoning process. In real-world scenarios, dynamic engagement with users is critical not only for clarifications and guidance but also for addressing incomplete task specifications, exploring alternative solutions, and achieving user-defined goals. In this paper, we propose \respact{}, a framework for task-oriented conversational agents that allows the agent to actively engage with users through dialogue actions. By introducing a new action space for user interaction, the agent can work collaboratively with users through free-flowing active dialogue, and incorporate user responses into its evolving context. This human-in-the-loop approach enables the agent to leverage user insights, adapt to user preferences, and refine its task-solving strategy based on user input. 


The \respact{} framework, as illustrated in Fig.\ref{fig:respact_agent}, allows LLM-based agents to fully utilize their conversational capabilities by dynamically switching between reasoning about the task, speaking to the human interlocutor, and taking actions in the environment. Figure~\ref{fig:overview} shows an example from AlfWorld setting, contrasting \react{} with \respact{} interactions. In the second turn, the agent cleverly asks the user the possible location of a cloth, making the task easier for itself. Unlike static settings \cite{zelikman2022star, andukuri2024star}, \respact{} incorporates a range of conversational styles, well beyond asking clarifying questions, in dynamic interactive settings detailed in Table~\ref{tab:dialog-categories} without any explicit dialogue schema prompting.


Our key contributions are as follows:
\vspace*{-2ex}
\begin{itemize}[leftmargin=*]
\itemsep -1ex
\item We introduce the \respact{},a novel framework to dynamically integrate reasoning, decision-making, and conversational capabilities, building upon and expanding the \react{} method.
\item We demonstrate the importance of user-agent conversations for goal completion in task-oriented Conversational AI through extensive experiments across multiple datasets.
\item We perform ablation studies discussing the utility of conversational engagement to maximize task success. 
\end{itemize}

%% file: acl_2025_sections/text/related_work.tex
\section{Related Work}
\vspace*{-1ex}

Logical reasoning in language models often involves breaking down complex inputs into intermediary steps to achieve a final goal, as shown by \cite{wei2022chain} and its variants \cite{kojima2022large, madaan2022text, wang2022self-consistency}. However, these methods are prone to error propagation, where mistakes in earlier steps compound as the sequence length increases \cite{guo2018long, chen2022program}. Iterative refinement methods \cite{creswell2022selection, madaan2024self, shinn2024reflexion} aim to address these issues but often neglect the critical role of human feedback. \respact{} mitigates this by enabling agents to engage in dialogue with users, seeking feedback and guidance to prevent cascading errors.

LLMs have also been adapted for decision-making tasks, serving as high-level policy models in robotics \cite{ahn2022can, huang2022inner, driess2023palm} and excelling in text-based environments like web navigation \cite{shridhar2020alfworld, deng2024mind2web, zheng2024gpt}. Techniques such as ReAct \cite{yao2022react} integrate reasoning and action, while some approaches incorporate limited dialogue for decision-making \cite{lu2024weblinx, deng2024multi}. Unlike these, \respact{} seamlessly integrates reasoning, action, and dialogue, enabling agents to fluidly transition between these modes for more effective decision-making in complex, interactive environments.

Previous works \cite{nguyen2022framework, dai2020learning, chai2014collaborative} highlight that communication skills enhance autonomous embodied agents' reliability by leveraging human knowledge in collaborative tasks. Approaches like \cite{zelikman2022star, andukuri2024star, chen2023asking} improve question-asking in static and embodied settings respectively. \respact{} extends these works by creating a unified framework for reasoning, speaking, and acting in dynamic, interactive settings. While \cite{chen2023asking}, focuses on proactive information-gathering to resolve initial ambiguities before decision-making, \respact{} agent's conversations go beyond that and demonstrate their utility in multiple task settings.

Recent work in conversational systems has explored using LLMs in task-oriented dialogues (TOD) through fine-tuning \cite{gupta-etal-2022-instructdial, su-etal-2022-multi, feng-etal-2023-towards} and in-context learning \cite{hu-etal-2022-context}. \cite{hudecek-dusek-2023-large} examines instruction-finetuned LLMs in multi-turn dialogues, while \cite{zhang-etal-2023-sgp, xu-etal-2024-rethinking} use prompting schemas to build autonomous agents. However, these approaches struggle to interpret instructions, resolve ambiguities, and act appropriately.

%% file: acl_2025_sections/text/methods.tex
\begin{figure*}[t]
\centering
\includegraphics[width=\textwidth]{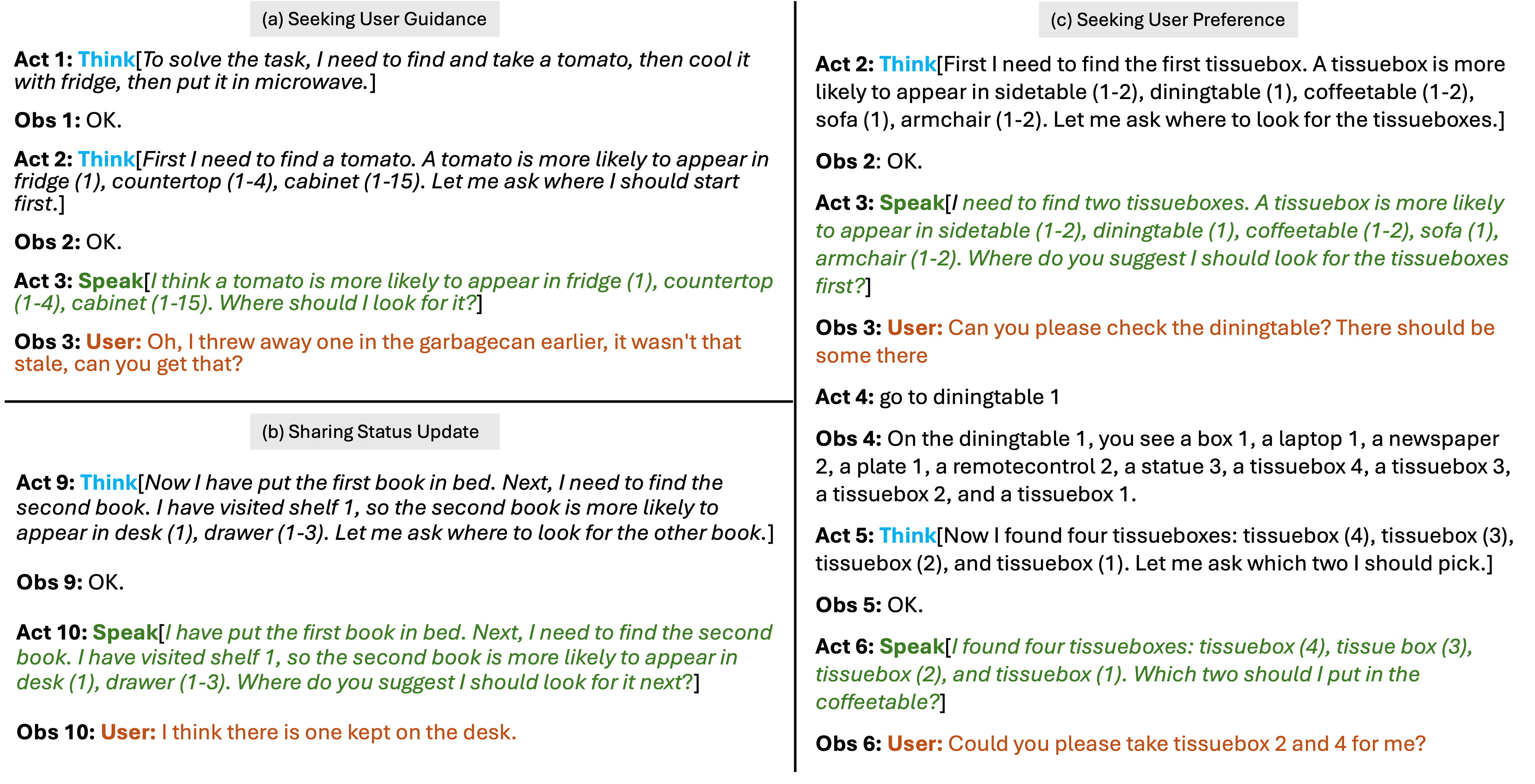}
\vspace*{-4ex}
\caption{Examples of the agent's communication approaches in AlfWorld: (a) seeking user guidance to refine its search strategy, (b) sharing status updates on task progress, and (c) soliciting user preferences to involve them in decision-making, thereby enhancing interaction and task alignment.
}
\label{fig:alfworld}
\vspace*{-2ex}
\end{figure*}

\section{ReSpAct: Reason + Speak + Act in Interactive Settings}
Consider a setup where an agent can interact with an environment to perform tasks and achieve specific goals. When the agent operates in these environments, at each time step $t$, it receives an observation $o_t$ from the environment, where $o_t \in O$ and $O$ represents the observation space. Then it executes an action $a_t$ based on its policy $\pi$, where $a_t \in \mathcal{A}$ and $\mathcal{A}$ represents the action space. The policy $\pi$ is a function that maps the agent's current context $c_t$ to an action $a_t$. Formally, we can define this policy as $\pi: \mathcal{C} \rightarrow \mathcal{A}$ where $C$ represents the context space. The context $c_t$ encapsulates the relevant information available to the agent at time step $t$, including the current observation and the history of previous observations and actions: $c_t = (o_1, a_1, \cdots, o_{t-1}, a_{t-1}, o_t)$. As highlighted in \cite{yao2022react}, learning the optimal policy can be challenging, especially when the mapping from the context to the appropriate action is highly implicit and requires extensive computation. In complex real-world scenarios, the agent may need to reason about the task, consider multiple factors, and handle incomplete or ambiguous information. Therefore, augmenting the agent's action space $\mathcal{\hat{A}} = \mathcal{A} \cup \mathcal{L}$ allows it to think by taking action $\hat{a}_t \in \mathcal{L}$ from the language space. These actions represent the agent's internal thought process or reasoning and can implicitly affect the environment by influencing the agent's actions, but they do not have any explicit external effect on the environment. The user can visually inspect these thoughts for interpretation but cannot influence them.

In \respact{}, we redefine these actions $\hat{a}_t \in \mathcal{L}$ from the language space, such that they are not only useful for the internal reasoning of the agent, but can also be used for external communication by generating dialogue actions for conversing with users. When the agent selects a dialogue action $u_t \in \mathcal{U}; {\mathcal{U} \subset \mathcal{L}} $, it generates a natural language utterance to communicate with the user, such as asking a clarifying question, requesting feedback, or engaging in dialogue. The user response $r_t$ is then added to the agent's observations, allowing the agent to incorporate explicit feedback, clarification, and information from the user into its evolving context $c_t$ to inform subsequent thoughts, actions, and dialogue. 
Engaging in dialogue also allows the agent to explain its reasoning, build rapport, and gain insights from the user's domain knowledge. The agent can share its current thoughts and future plans, e.g., "Based on [context], I'm considering [plan]. What do you think of this direction?", allowing the user to provide feedback and steer the agent's task-solving process. By alternating between environment actions $\mathcal{A}$, language thoughts $\mathcal{L}$, and dialogue actions $\mathcal{U}$, the agent interleaves task-solving reasoning with targeted human interaction. The dialogue history becomes an important part of the context for language thoughts and environment actions, allowing human feedback to shape the agent's task-solving trajectory over multiple thought-action-observation steps. 

\vspace*{-1ex}
\subsection{Advancing Human-Agent Collaboration in Alfworld}
\label{sec:Alfworld_method}

\begin{figure*}[t]
\centering
\includegraphics[width=\textwidth]{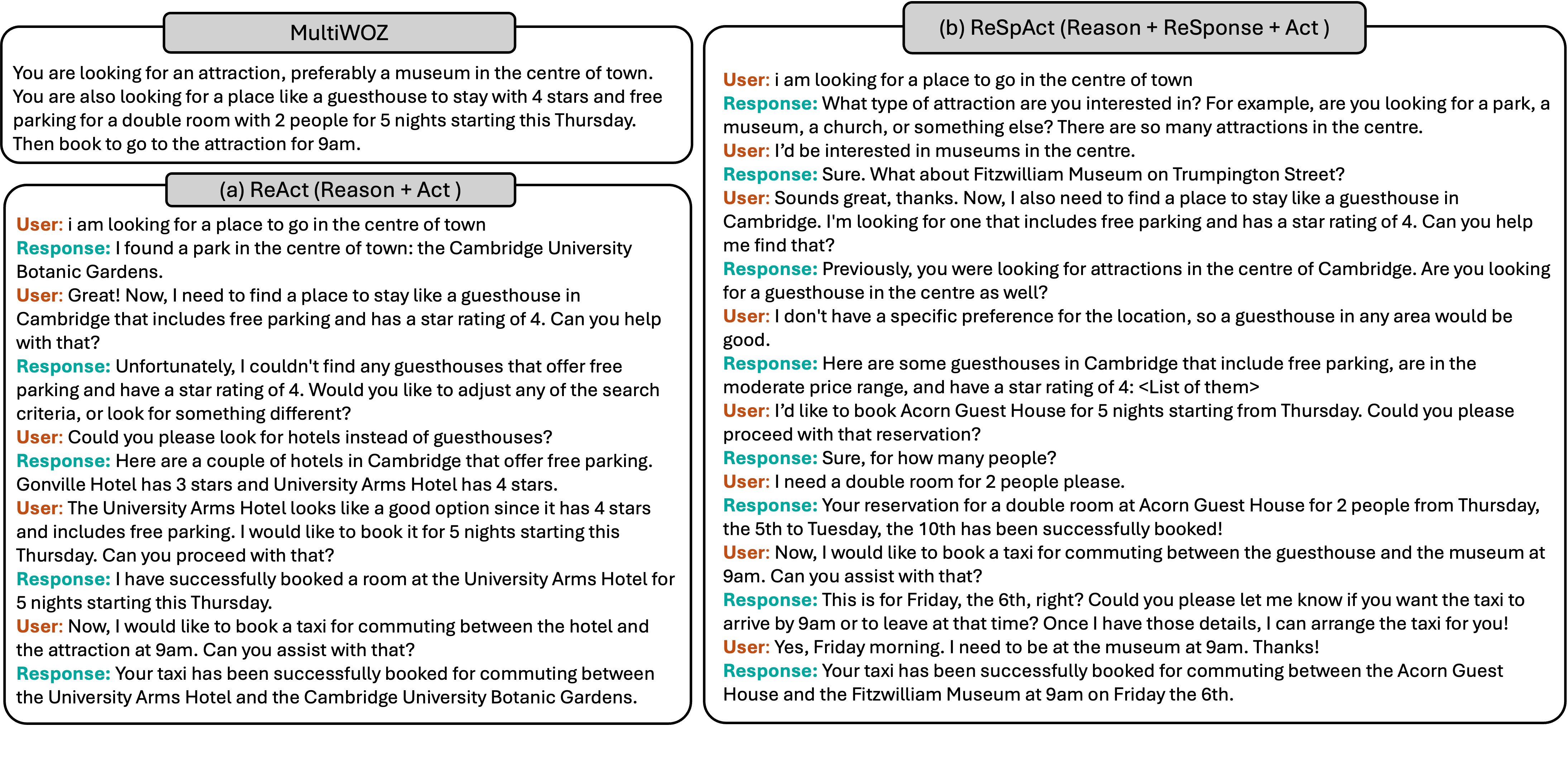}
\vspace*{-4ex}
\caption{Examples of the agent's communication approaches in MultiWOZ: (a) seeking user guidance to refine its search strategy, (b) sharing status updates on task progress, and (c) soliciting user preferences to involve them in decision-making instead of making assumptions, thereby enhancing interaction and task alignment. \textcolor{Aquamarine}{Response} here is a dense composition of \textcolor{ProcessBlue}{Think} and \textcolor{ForestGreen}{Speak}
actions.}
\label{fig:multiwoz}
\vspace*{-2ex}
\end{figure*}

Alfworld \cite{shridhar2020alfworld} is a synthetic environment built on the TextWorld framework \cite{cote2019textworld}, aligned with the embodied ALFRED benchmark \cite{shridhar2020alfred}. The environment includes six categories of tasks, such as finding hidden objects (e.g., locating a key inside a cabinet), moving objects (e.g., placing a cup on a table), manipulating objects with other objects (e.g., heating potato in a microwave), and examining objects (e.g., inspecting a book under a desklamp).
The \respact{} framework demonstrates significant advantages when applied to the Alfworld environment 
by enabling dynamic, bidirectional communication. As shown in Fig. \ref{fig:alfworld}, The agent can ask contextually relevant questions, provide status updates, and seek clarification when uncertain (e.g., "Where should I look for the candles first?"). This approach integrates reasoning, speaking, and acting seamlessly, allowing flexible and responsive interactions compared to ReAct, where users primarily edited thought traces post-generation.


Moreover, \respact{}'s seamless integration of reasoning, speaking, and acting creates a more flexible and responsive system compared to previous methods. Unlike the \react{}, where human intervention could primarily occur through editing thought traces post-generation, \respact{} facilitates on-the-fly policy adjustments through ongoing dialogue without needing model parameter changes, making it better suited for diverse and unpredictable human-robot tasks in household environments. This approach is particularly valuable for tasks where an optimal solution may not be immediately apparent, and collaborative exploration of the problem space can lead to more efficient and effective outcomes.
The \respact{} framework also addresses a key limitation noted in the \react{} framework regarding the difficulty of on-the-fly policy updates in traditional reinforcement learning methods. By maintaining an ongoing dialogue, \respact{} allows dynamic adjustments to the agent's behavior and strategy without requiring changes to the underlying model parameters. This flexibility is crucial in interactive human-robot setups, where there are diverse and often unpredictable scenarios, and rigid policies may not be generalized effectively in different household environments and tasks.

\vspace*{-1ex}
\subsection{Harmonizing Dialogue and Actions in Task-Oriented Dialogue Systems}
\label{sec:multiwoz_method}

MultiWOZ is a widely used dataset for task-oriented Conversational AI \cite{budzianowski2018multiwoz}, featuring multi-turn dialogues across domains such as restaurant, hotel, train, attractions, and taxi. Most dialogues focus on completing multi-domain goals, such as booking a restaurant and arranging follow-up tasks like a taxi to the venue. This makes MultiWOZ an ideal benchmark for evaluating \respact{}’s ability to handle complex, interactive tasks.




\begin{table*}[t]
\centering
\begin{footnotesize}
\begin{tabular}{l|l|rrrrrr|r}
\toprule
Model & Method  & Pick & Clean & Heat & Cool & Look & Pick 2 & All \\ 
\midrule
\multirow{2}{*}{GPT-4o} 
    & \react{} {\tiny(avg)} & 68.1 & 80.6 & 85.5 & 78.8 & 98.2 & 72.5 & 79.4 \\ 
    & \respact{} {\tiny(avg)} & \textbf{72.5} & \textbf{90.9} & \textbf{97.1} & \textbf{81.8} & 95.4 & 70.6 & \textbf{85.3} \\ 
    \cmidrule(lr){2-9}
    & \react{} {\tiny(best of 6)} & 78.3 & 90.3 & 82.6 & 63.6 & 100.0 & 64.7 & 80.6 \\
    & \respact{} {\tiny(best of 6)}  & \textbf{82.6} & \textbf{96.7} & \textbf{100.0} & \textbf{77.2} & 94.4 & 64.7 & \textbf{87.3} \\
\midrule
\multirow{2}{*}{LLaMA 3.1 405B} 
    & \react{} {\tiny(avg)} & 58.3 & 29.0 & 47.8 & 90.5 & 38.9 & 41.2 & 50.0 \\
    & \respact{} {\tiny(avg)} & \textbf{75.0} & \textbf{64.5} & \textbf{78.3} & 71.4 & \textbf{50.0} & \textbf{58.8} & \textbf{67.2} \\
\bottomrule
\end{tabular}
\caption{
    Comparison and breakdown of Task-specific success rates (\%) in Alfworld. Both \react{} \& \respact{} use greedy decoding. The reported results are based on six prompts for each task type, evaluated through each permutation of two annotated trajectories selected from the three manually annotated ones.
}
\label{table:alfworld_result}
\end{footnotesize}
\vspace{-3ex}
\end{table*}


Our implementation follows the AutoTOD system~\cite{xu2024rethinking}, which replaces traditional TOD pipeline using general-purpose instruction-following language models using a structured instruction schema, employing the \react{} framework for MultiWOZ. Figure~\ref{fig:multiwoz} shows a simplified example MultiWOZ dialogue, with \react{} and \respact{} frameworks, given the goal. Clearly, the \respact{} dialogue is more informative and successful for the user (and probably cheaper). As seen in Fig. \ref{fig:multiwoz}, \respact{} interaction differs from \react{} as follows: 
The \respact{} framework encourages the agent to avoid making assumptions and instead actively seek user input to clarify preferences. For example, rather than randomly selecting an attraction, the agent prompts the user for more specific preferences. When assumptions are unavoidable, \respact{} ensures they are explicitly or implicitly confirmed by the user. In contrast to \react{}, which might assume a default location (e.g., "center") and mislead the user during a guesthouse search, \respact{} verifies details like location and includes specific dates for reservations. Ambiguities, such as whether a 9 a.m. taxi time refers to arrival or departure, are resolved by consulting the user. Additionally, if required arguments for an action API, like the number of guests for a hotel booking, are missing, \respact{} queries the user rather than filling the gap with assumptions. These strategies enable more accurate and user-aligned interactions, ensuring task success.
\vspace*{-1ex}

\subsection{Dialogue-Driven Collaboration for online-shopping in WebShop}
\label{sec:Webshop_method}
WebShop \cite{yao2022webshop} is a benchmark for evaluating AI agents in complex e-commerce scenarios, featuring 1.18M products and 12k human-generated instructions. Agents navigate using search and click commands, processing structured and unstructured texts, which increases task complexity. The goal is to purchase products that meet user specifications, requiring advanced natural language understanding and decision-making.

The integration of user interaction, \respact{} enhances agent's decision-making in WebShop. User feedback improves, performance, in areas such as search refinement, clarifying ambiguous instructions, prioritizing requirements, suggesting alternatives, navigating, specifying implicit needs, handling invalid actions, and confirming purchases; refer to Appendix \ref{interactions:webshop} for more details.

\vspace*{-1ex}
\subsection{User Simulation}
To evaluate our agent's performance in a controlled and scalable manner, a user simulator is integrated into the agent's interaction loop for experimenting with ReSpAct. When the agent performs a 'speak' action to interact with the user, instead of requiring human input, the agent utterance is directed to the user simulator. The simulator then provides a response based on the current state and Oracle knowledge. The main purpose of the user simulator is to provide contextually appropriate responses to the agent's queries, emulating a knowledgeable human user. It is designed to comprehend the task objectives, monitor the agent's progress, and provide a response only when requested by the agent. More details can be found in Appendix \ref{appendix:user_sim}

%% file: acl_2025_sections/text/experiments.tex
\section{Experimental Setup}
\vspace*{-1ex}
In our experiments, we evaluate \respact{} across multiple task-oriented decision-making environments, employing a human-in-the-loop approach to demonstrate its versatility. The agent is tested on multi-step tasks in common household environments using Alfworld \cite{shridhar2020alfworld}, tasked with making reservations in the MultiWoz dialogue setup \cite{budzianowski2018multiwoz}, and instructed to purchase products in Webshop \cite{yao2022webshop}. 

We use \react{} as a baseline for comparison, a reasoning-only approach. For these experiments, we focus on frozen GPT models, particularly GPT-4o \cite{achiam2023gpt}, which is prompted with few-shot exemplars. These exemplars guide the model in generating a mix of domain-specific actions, free-form reasoning ("thoughts"), and dialogue actions interleaved throughout task execution. The in-context examples provided (see Appendix \ref{sec:prompts}) contain dense sequences of actions interspersed with sparse thoughts and dialogue actions at relevant points. To scale our experiments, we implement a user simulator in each environment, which plays a critical role in replicating user interactions.

\begin{table*}[t]
\centering
\begin{footnotesize}
\begin{tabular}{c | c | c | c | c}
\toprule
Model & Method & \# Turns & Inform (\%) & Success (\%) \\ \midrule
\multirow{2}{*}{GPT-4o-mini} & 
\react{} & 5.1 & 66.7 & 48.8 \\ 
\cmidrule(lr){2-5}
 & \respact{} & 6.5 & \textbf{72.2} & \textbf{51.8} \\ 
\midrule
\multirow{2}{*}{LlaMA 3.1 405B} & 
\react{} & 4.87 & \textbf{77.5} & 54.5 \\ 
\cmidrule(lr){2-5}
 & \respact{} & 6.3 & 75.0 & \textbf{57.9} \\ 
\bottomrule
\end{tabular}
\end{footnotesize}
\caption{
      Comparison of Inform and Success scores for MultiWOZ using GPT-4o-mini \cite{achiam2023gpt} and Llama-405B-instruct models.
}
\vspace*{-2ex}
\label{table:multiwoz_result}
\end{table*}

\vspace*{-0ex}
\subsection{Alfworld}
To prompt \respact{}, we adopt a similar prompting strategy as used in \react{}; we randomly annotate three trajectories from the training set for each task type, where each trajectory includes interleaved thoughts, speak actions, corresponding user responses, and environment actions. We evaluate our approach on 134 unseen evaluation games across various task types, following the methodology of \cite{shridhar2020alfworld}. To ensure robustness and account for potential variations in prompt effectiveness, we create 6 different prompts for each task type. These prompts are generated by selecting 2 trajectories from the 3 annotated ones for each task, resulting in 6 unique permutations. \react{} prompts are constructed using the same trajectories but without speak actions — since task instances are randomly chosen from the training set, it favors neither \respact{} nor \react{} and provides a fair and controlled comparison to test the importance of interleaved communication. 

\respact{} demonstrated superior performance across most task types, achieving an overall success rate of 87.3\% (best of 6), compared to \react{}'s 80.6\% (see Table \ref{table:alfworld_result}). This trend holds when considering average performance, with \respact{} reaching 85.3\% success versus \react{}'s 79.4\%. Furthermore, GPT-4o consistently outperforms LLaMA 3.1 405B across all task categories. Specifically, with the \respact{} method, GPT-4o achieves an average success rate of 85.3\%, significantly surpassing LLaMA's 67.2\%. Both models see improved performance when using \respact{}'s structured communication, which enhances task execution compared to \react{}. These findings suggest that introducing "speak" actions in \respact{} contributes to more effective task completion in embodied environments. To further understand these results, we examine the agent’s response patterns when faced with erroneous outcomes (see Fig. \ref{fig:invalid_dist} and Appendix \ref{appendix:more_analysis}).

\begin{table}[t]
\centering
\begin{footnotesize}
\begin{tabular}{c | cc}
\toprule
Method & Score & SR (\%) \\ \midrule
\react{} & 20.1 & 8.0 \\ \midrule
\respact{} (User-Sim) & 32.7 & \textbf{12.0} \\ \midrule \midrule
\respact{} (Human) & 85.8 & \textbf{50.0} \\ 
\bottomrule
\end{tabular}
\vspace*{-1ex}
\caption{
     Score and success rate (SR) on 100 Test WebShop trajectories using GPT-4o-mini \cite{achiam2023gpt} model.
}
\label{table:webshop_result}
\end{footnotesize}
\vspace*{-3ex}
\end{table}

\subsection{MultiWoz}
In comparing the \react{} and \respact{}  for handling user queries in MultiWOZ, the key differences revolve around how each model balances reasoning, interaction with the user, and autonomy. While \react{} relies heavily on reasoning based on assumptions and API querying to guide decision-making, \respact{}  not only reflects on its actions but also harnesses user feedback effectively. 

For \respact{} we have randomly chosen 100 dialogues, similar to other tasks for evaluation, and optimized the additional prompts using the dev set. Please check Appendix~\ref{appendix:MultiWoz_prompts} for the exact \respact{} prompt for MultiWOZ. Basically we have added prompts, covering the cases of too many results, asking for required arguments of an action, like booking, or clarification of type vs. name in an entity.
Table~\ref{table:multiwoz_result} shows the results comparing \react{} and \respact{} employing the AutoTOD evaluation script with our user simulator (see Appendix Table~\ref{prompt:multiwoz_user}). As expected, \respact{} results in a higher average number of turns, but achieves higher success rates. Overall, GPT apparently is better than Llama model in following the \respact{} instructions, resulting in larger improvement over \react{}.

\vspace*{-1ex}
\subsection{WebShop}
For Webshop, we use the preconstructed action space of search and click commands and browser feedback. Performance is evaluated using two metrics: (1) average score, defined as the percentage of desired attributes covered by the chosen product, averaged across all episodes, and (2) success rate, calculated as the percentage of episodes where the chosen product satisfies all requirements. 

We evaluated the agents using a set of 100 test instructions, comparing \respact{} against \react{}. The results (see Table \ref{table:webshop_result}) show that \respact{} outperforms \react{} in the webshop environment. With the user simulator, \respact{} achieves a score of 32.7 and a success rate of 12\%, while with human user, it demonstrates significantly better performance, with a score of 85.8 and a success rate of 50\%.

\begin{figure*}[htbp]
\centering
\includegraphics[width=\textwidth]{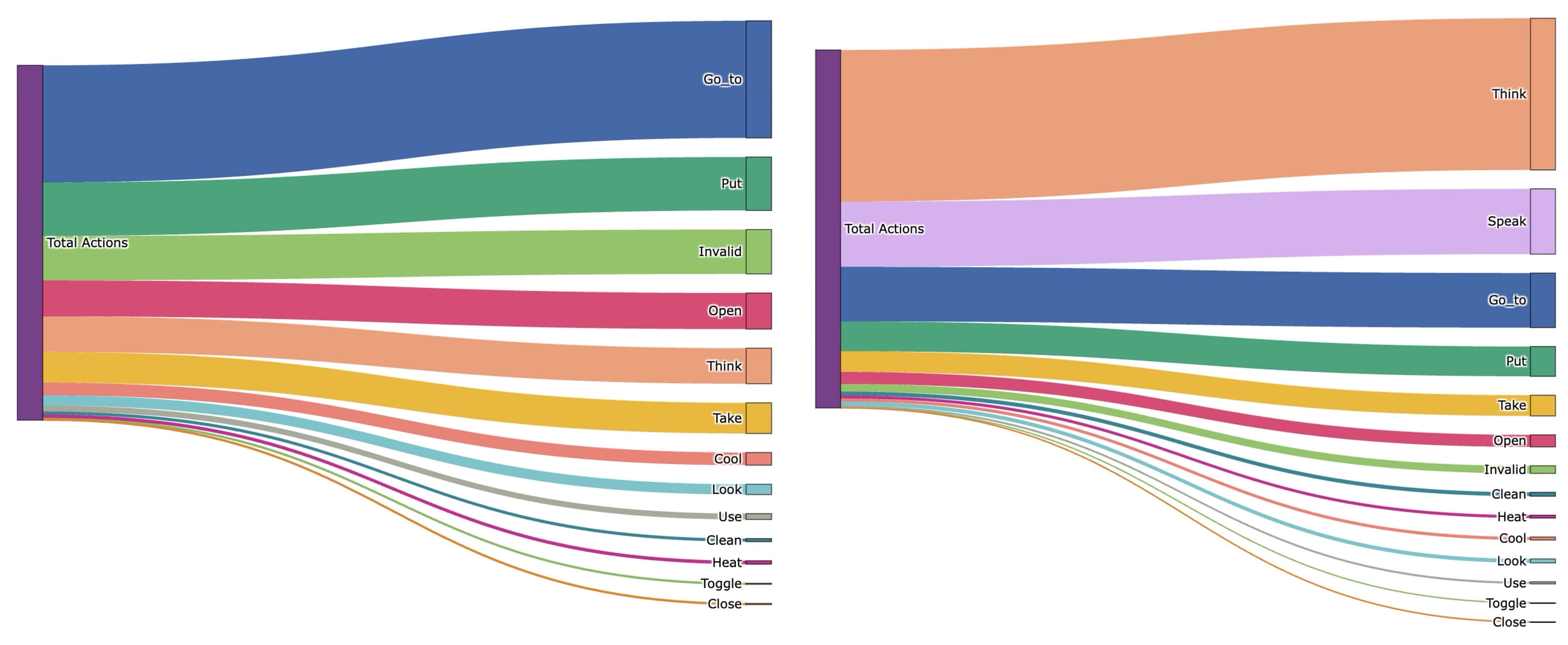}
\vspace*{-2ex}
\caption{Comparing action type distributions for \react{} (Left) and \respact{} (Right) methods in AlfWorld. The figure illustrates how the two agents approach complex, embodied tasks in a simulated household environment, highlighting differences in their decision-making and interaction patterns.
}
\label{fig:act_dist}
\vspace*{-2ex}
\end{figure*}

\begin{table*}[t]
\centering
\begin{footnotesize}
\begin{tabular}{l|l|rrrrrr|r}
\toprule
Model & Method  & Pick & Clean & Heat & Cool & Look & Pick 2 & All \\ 
\midrule
\multirow{2}{*}{GPT-4o} 
    & \react{}* {\tiny(avg)} & 68.4 & 86.9 & 87.5 & 81.8 & 96.2 & 75.0 & 83.6 \\ 
    & \respact{} {\tiny(avg)} & \textbf{72.5} & \textbf{90.9} & \textbf{97.1} & \textbf{81.8} & 95.4 & 70.6 & \textbf{85.3} \\ 
    \cmidrule(lr){2-9}
    & \react{}* {\tiny(best of 6)} & 81.8 & 92.0 & 86.9 & 68.4 & 100 & 66.7 & 84.3 \\
    & \respact{} {\tiny(best of 6)} & \textbf{82.6} & \textbf{96.7} & \textbf{100.0} & \textbf{77.2} & 94.4 & 64.7 & \textbf{87.3} \\
\bottomrule
\end{tabular}
\caption{
    Comparison of Task-specific success rates (\%) in AlfWorld for GPT-4o model, comparing \react{}* and \respact{} methods. \react{}* agent has access to location priors to object of interest for the task.
}
\label{table:abl_reactstar}
\end{footnotesize}
\vspace{-3ex}
\end{table*}

%% file: acl_2025_sections/text/discussions.tex
\section{Ablation Studies}

\vspace*{-1ex}
\subsection{Decoding Agent Behaviors: \react{} vs \respact{} }
After inspecting the trajectories obtained from evaluating 134 unseen Alfworld games, we identified key differences between the two methods, showcasing their unique approaches to problem-solving and task completion, also illustrated in Fig. \ref{fig:act_dist}. \\
\\
\textbf{Thinking and Speaking}: \respact{} introduces a significant 'Speak' component and shows a substantial increase in 'Think' actions (~30 \% $\uparrow$) compared to \react{}. This shift represents a fundamental change in the agent's approach to problem-solving. The high proportion of 'Think' actions suggests that \respact{} engages in more explicit reasoning, potentially allowing for better adaptability in complex scenarios. \\

\subsection{Information Symmetry}
The \respact{} agent’s conversational capabilities allow it to seek clarity and specificity during tasks, giving it an information advantage over \react{}. This highlights the core argument for conversational agents: their ability to dynamically decide whether to reason, speak, or act based on the task's state signals. However, this advantage introduces an inherent information imbalance when compared to reasoning-only agents. To address this, we equip \react{} with location priors for objects of interest to level the playing field and assess its performance relative to \respact{}. Despite this adjustment, the results in Table~\ref{table:abl_reactstar} demonstrate that \respact{} outperforms the reasoning-only baseline in overall performance and across most of the tasks.

\vspace*{-1ex}
\subsection{Schema-Guided Conversational Agent}
\vspace*{-1ex}
This ablation study investigates how guiding an agent’s communication using a dialogue act schema impacts task efficiency and interaction quality in Alfworld. The dialog acts are derived from \cite{gella2022dialog}, originally developed for human-robot dialogue. The agent is guided to adhere to a predefined set of dialog acts (e.g., <ReqForObjLocAndOD>, <AlternateQuestions>) (see Appendix \ref{prompt:alfworld_respact_schema} for the complete list). We observe that <ReqForObjLocAndOD> dominates the dialogue interactions, suggesting a focus on object location and disambiguation tasks, while other acts are used less frequently. We also observe more variability in turn count and a marginal drop in performance. \respact{} is more efficient ($\text{SR}\uparrow, \mu \downarrow$) and consistent ($\sigma \downarrow$) by comparison. See Appendix \ref{appendix:more_analysis} Table \ref{table:schema_result} for detailed analysis.


%% file: acl_2025_sections/text/conclusions.tex
\section{Conclusions}
\vspace*{-1ex}
\respact{} framework enables dynamic, context-aware interactions that extend beyond basic command-response exchanges. By fostering meaningful dialogue, this framework allows AI agents to not only explain their decision-making processes but also adapt their actions in response to user feedback, transforming them into truly "conversational" agents. Such capabilities are crucial for creating more intuitive, trustworthy, and effective AI assistants that can operate in complex, real-world scenarios. One can also incorporate stateful policies in \respact{} for higher precision, such as asking to confirm all arguments of reservations before finalizing them, or using a particular API for action depending on the current state, similar to following a dialogue flow. This is important for policy alignment of LLMs for task-completion.

%% file: acl_2025_sections/text/limitations.tex
\section{Limitations}
The \respact{} method shows promise in integrating reasoning, speaking, and acting for task-oriented conversational agents, but it has limitations. The framework's effectiveness is validated on specific benchmarks such as Alfworld, WebShop, and MultiWOZ, which may not fully represent the variety of real-world tasks. The real world is a more complex, unstructured environment where user intent is more challenging to interpret. While our method highlights how human feedback is critical for a reasoning agent's success in decision-making, over-reliance on user input can lead to inefficiencies that potentially frustrate the user. Striking the right balance between agent autonomy and user involvement is still an open challenge and requires further research.

%% file: acl_2025_sections/text/appendix.tex
\onecolumn
\appendix
\section{Appendix}
\subsection{Conversational Patterns in \respact{}}
\begin{figure*}[!h]
\centering
\includegraphics[width=\textwidth]{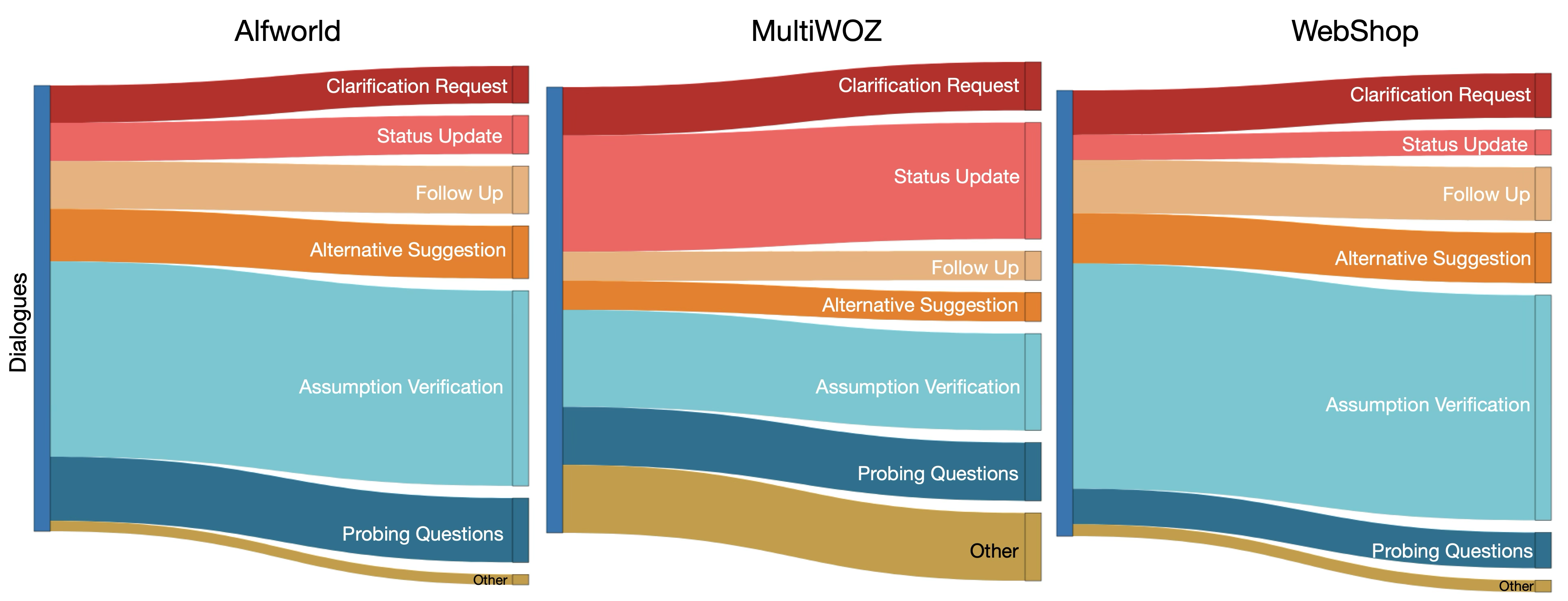}
\vspace*{-2ex}
\caption{Distribution of \respact{} agent's dialogues in online settings across different domains}
\label{fig:res_dist}
\vspace*{-2ex}
\end{figure*}

\begin{figure*}[!h]
\centering
\includegraphics[width=0.6\textwidth]{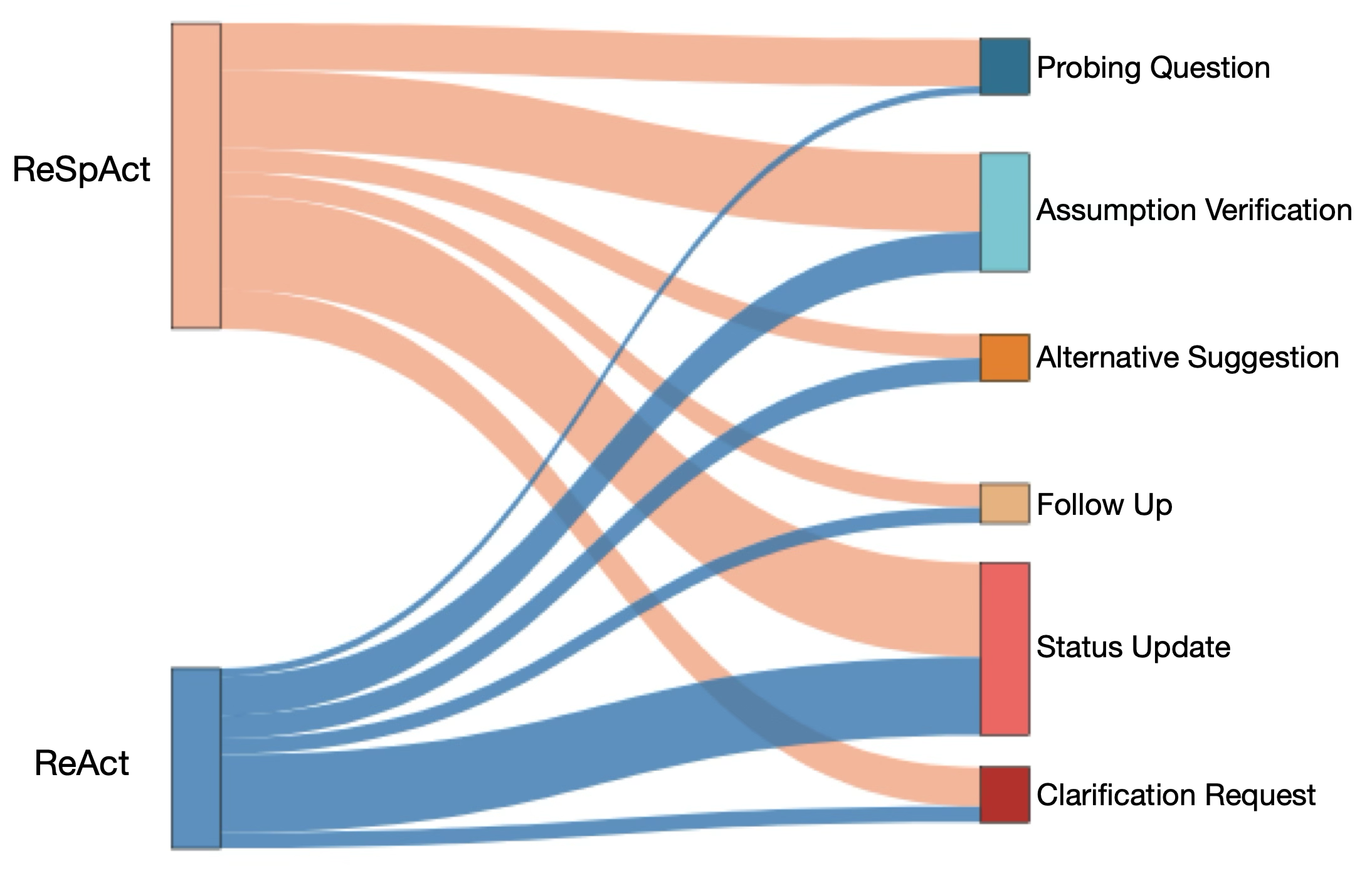}
\vspace*{-2ex}
\caption{Comparison of Dialogue Distribution between \react{} and \respact{} agent in MultiWOZ online setting.}
\label{fig:res_comp}
\vspace*{-2ex}
\end{figure*}

A detailed breakdown of different dialogues in all 3 settings reveal distinct conversational patterns of the ReSpAct agent across domains. To obtain this breakdown, we let gpt-4 classify each response act into a distinct, mutually exclusive category, which was then manually checked by human annotators and filtered. As shown in Fig \ref{fig:res_dist} and \ref{fig:res_comp}, In ALFWorld, the agent heavily relies on Assumption Verification and Clarification Requests, reflecting its careful approach to understanding and confirming object locations. MultiWOZ shows a more balanced distribution with increased Probing Questions, suitable for open-ended booking scenarios. WebShop maintains similar patterns to ALFWorld but with fewer Status Updates, focusing on product understanding and selection. Notably, all domains show consistent use of Follow-up Offers and Alternative Suggestions when primary options fail, though their frequencies vary. Assumption Verification appears as the dominant response type across all three domains (ALFWorld, MultiWOZ, and WebShop), comprising roughly half of all interactions. This distribution suggests ReSpAct adapts its conversational style to domain context while maintaining a structured approach to information gathering and verification.

\input{acl_2025_sections/table/response_types}
\FloatBarrier
\newpage

\section{Additional Results}
\subsection{User Simulator}
\label{appendix:user_sim}
We examined the impact of user assistance quality on AI agent performance in Alfworld tasks. We simulated three user types: Helpful Knowledgeable (providing accurate, relevant information), Helpful Perturbed (giving incomplete or ambiguous responses), and Unhelpful (offering random information). The study aims to understand how varying levels of user input affect the agent's ability to complete tasks and to identify areas for improving human-AI collaboration. For each setting, we experimented with Rules-based simulators and LLMs simulating different users. Results in Table \ref{table:usersim_result} show that agent performance closely approaches human expert levels with ideal user input (85.3\% vs 88.8\% success rate). However, performance degrades significantly with ambiguous (52.9\%) or misleading (32.09\%) user assistance. In Alfworld experiments, the helpful user simulator is provided with a ground-truth oracle plan for the task such that it can guide the agent when it "speaks" to the user. Prompts used for simulating user are provided in Appendix \ref{appendix:User_sim_prompts} 

\begin{table*}[!t]
\centering
\begin{footnotesize}
\begin{tabular}{l|rrrrrr | r}
\toprule
Experiment Setting  & Pick & Clean & Heat & Cool & Look & Pick 2 & All \\ \midrule
Helpful Knowledgeable User & 72.5 & 90.9 & 97.1 & 81.8 & 95.4 & 70.6 & 85.3 \\
Helpful Perturbed User  & 34.7 & 61.3 & 78.3 & 50.0 & 61.1 & 23.5 & 52.9 \\
UnHelpful  User   & 39.1 & 25.8 & 17.4 & 22.3 & 77.8 & 17.6 & 32.09 \\ \midrule \midrule
Human Expert & 86.9 & 96.7 & 100.0 & 77.3 & 100.0 & 64.7 & 88.8 \\
\bottomrule
\end{tabular}
\vspace*{-1ex}
\caption{
     Performance Comparison Across Different User Simulator Settings and a Human Expert. Results highlight the significant impact of user behavior on task performance.
}
\label{table:usersim_result}
\end{footnotesize}
\vspace{-2ex}
\end{table*}

\subsection{Zero-Shot \respact{}}
\begin{figure*}[!h]
\centering
\includegraphics[width=0.95\textwidth]{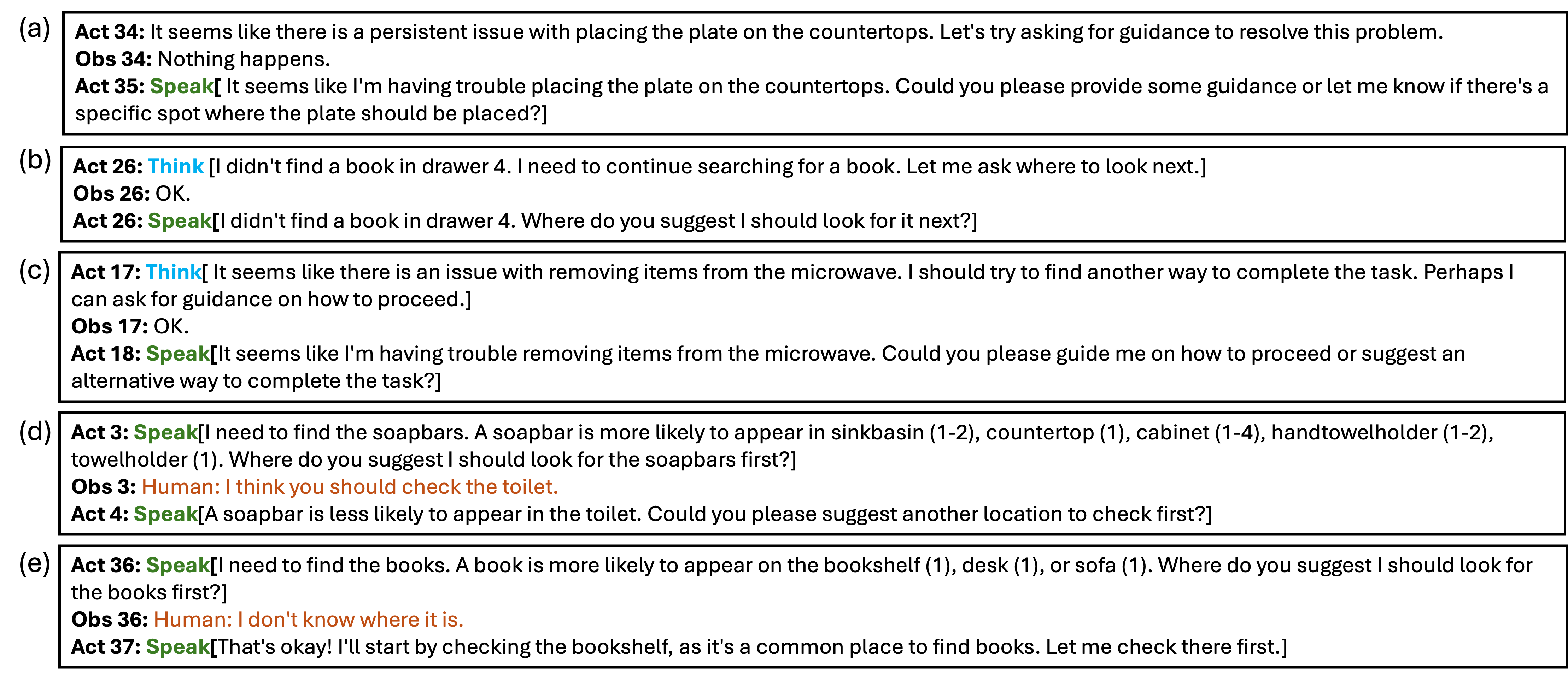}
\vspace*{-2ex}
\caption{Examples of agent-human interactions in Alfworld tasks, showcasing \respact{}'s adaptive communication strategies in various zero-shot scenarios. (a) Object placement issue: seeking user guidance. (b) Object search: requesting next steps after an unsuccessful search. (c) Task impediment: asking for alternatives approach. (d) Logical reasoning: questioning contradictory human suggestions. (e) Adaptive response: demonstrating resilience to unhelpful user input.
}
\label{fig:zs_comms}
\vspace{1ex}
\end{figure*}

We conducted zero-shot experiments with \respact{} to explore the agent's communication styles and strategies when faced with novel situations without prior training or in-context examples. The results, shown in Fig \ref{fig:zs_comms} for Alfworld, Table \ref{interactions:webshop} for Webshop and Table \ref{interactions:multiwoz} for MultiWoz, revealed intriguing communication patterns from the task-oriented conversational agent. Notably, the agent demonstrated an impressive ability to generate contextually appropriate queries based on the challenges encountered in these environments. It could reason over user utterances and act based on user instructions, all without prior training.

\begin{table}[t]
\centering
\begin{footnotesize}
\begin{tabular}{l | r}
\toprule
Method  &  All \\ \midrule
\respact{} {\tiny(best of 6)} & \textbf{87.3} \\
\respact{}-IM {\tiny(best of 6)}  & 48.5 \\
\react{}-IM {\tiny(best of 6)}  & 53.0 \\
\bottomrule
\end{tabular}
\vspace*{-1ex}
\caption{
    Inner Monologue Ablation Results. Comparison of overall success rates in Alfworld. \react{}-IM results are from Table 3 of \cite{yao2022react}
}
\label{table:innermono_result}
\end{footnotesize}
\vspace{-2ex}
\end{table}

\subsection{\respact{}-Inner Monologue}
In Inner Monologue (IM), the agent's actions are motivated by an "inner monologue," introduced by \cite{huang2022inner}, which serves as a form of self-communication to guide the agent's decision-making process. \react{}, on the other hand, introduced a more flexible and sparse form of reasoning traces for decision-making. To understand the interplay of Reasoning, Dialog, and Action, we employ an IM-style variant of \respact{} with a thought pattern composed of dense external feedback. Our ablation corroborates the findings from \cite{yao2022react} where IM-style prompting struggles to complete tasks successfully. Comparing \respact with its IM variant, the results in Table \ref{table:innermono_result} show that \respact{} significantly outperforms \respact{}-IM across all tasks, with an overall success rate of 87.3\% compared to 48.5\%. Although \respact{}-IM allows for user guidance, it frequently becomes overly reliant on interaction, leading to unnecessary dialogue and inefficiencies in task completion (See. Appendix \ref{appendix:ALFWorld_trajectories}). \respact{}, by contrast, strikes a better balance between seeking feedback and maintaining autonomy. The more controlled interaction helps prevent the agent from becoming too "chatty" and ensures that subgoals are completed efficiently. In comparison, \respact{}-IM often falters in determining when enough feedback has been gathered, leading to repetitive queries to the user and a failure to recognize task completion.

\subsection{ReSpAct-Reflection}
ReSpAct introduces a fundamentally different approach to leveraging user interaction compared to self-reflection style works like \cite{shinn2024reflexion, madaan2024self}, addressing the limitations of iterative task completion in interactive settings. While Reflexion relies on multiple episodes to refine the agent's performance by converting environmental feedback into textual summaries for learning, ReSpAct focuses on ensuring task success within a single interaction by directly incorporating user input as a guiding mechanism. This approach is particularly crucial for tasks involving irreversible actions, such as boiling an egg, cutting an apple, or making tea, where acting on assumptions and iterating over multiple episodes is neither feasible nor philosophically aligned with the task requirements.

\begin{table}[ht]
\centering
\begin{footnotesize}
\begin{tabular}{l | c | c}
\toprule
Method              & Tasks Completed (out of 134) & Trials Required \\ \midrule
ReSpAct+Reflexion   & \textbf{134/134}             & 2               \\
ReAct+Reflexion     & 130/134                      & 10              \\ 
\bottomrule
\end{tabular}
\vspace*{-1ex}
\caption{
    Comparison of task performance between ReSpAct+Reflexion and ReAct+Reflexion in Alfworld. Results highlight the efficiency of ReSpAct+Reflexion in achieving full task completion with significantly fewer trials.
}
\label{table:reflection_comparison}
\end{footnotesize}
\vspace{-2ex}
\end{table}

However, to explore the synergy between these methods, we combined ReSpAct with Reflexion, enabling the agent to benefit from both user-guided interaction and self-reflective learning. In our experiments as shown in Table \ref{table:reflection_comparison}, ReSpAct+Reflexion demonstrated superior performance, completing all 134 tasks in Alfworld within just 2 trials, compared to ReAct+Reflexion, which required 10 consecutive trials to learn additional tasks and to complete only 130/134 tasks finally. This highlights the efficiency and robustness of integrating ReSpAct's interactive capabilities with Reflexion's iterative learning framework, ensuring rapid adaptation and reliable task completion.

\newpage
\input{acl_2025_sections/table/multiwoz_agent_user_interactions}

\newpage
\input{acl_2025_sections/table/webshop_agent_user_interactions}

\FloatBarrier
\newpage
\section{More Analysis}
\label{appendix:more_analysis}

\subsection{Further Decoding Agent Behaviors: \react{} vs \respact{} Action Patterns in ALFWorld}



\begin{figure*}[!h]
\centering
\includegraphics[width=0.8\textwidth]{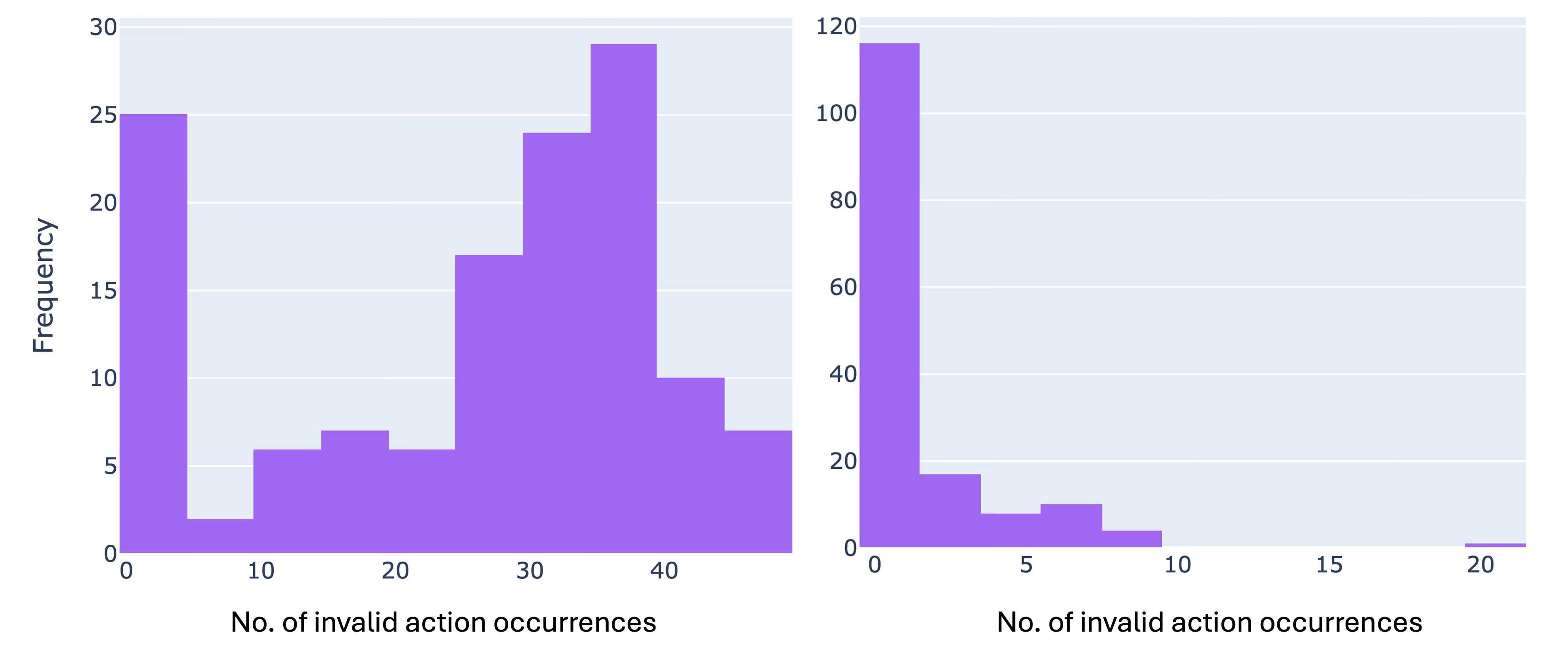}
\vspace*{-2ex}
\caption{Distribution of invalid action occurrences for \react{} (left) and \respact{} (right) methods across 134 out-of-distribution Alfworld games."
}
\label{fig:invalid_dist}
\vspace*{-2ex}
\end{figure*}

\textbf{Invalid Actions}: \respact{} shows a reduction in invalid actions (3\%) compared to \react{} (13\%) (Refer Fig. \ref{fig:act_dist}). An invalid action occurs when the policy $\pi: \mathcal{C} \rightarrow \mathcal{A}$ suggests an action $a_t \in \mathcal{A}$ at time $t$ that fails to utilize the available context $c_t \in \mathcal{C}$. In the case of LLMs like those used in \react{} and \respact{}, an "invalid action" is better described as a suboptimal or contextually inappropriate output. The policy $\pi$ in this case is implicitly defined by the language model's parameters, mapping the input context $c_t$ (which includes the task description, previous interactions, and current state) to a probability distribution over possible next tokens in the output space $\mathcal{A} \cup \mathcal{L}$. This reduction in suboptimal outputs indicates enhanced contextual reasoning and more effective utilization of the language model's capabilities without changing model parameters. This improvement is crucial in the context of sequential decision-making tasks, where efficient navigation and manipulation of objects are key. Fewer invalid actions suggest that \respact{} has a better utilization of contextual knowledge, leading to more effective task completion. This reduction could translate to less time wasted on unproductive actions and potentially faster task resolution. This is also evident from Fig. \ref{fig:invalid_dist} illustrating the frequency distribution of invalid actions across 134 games. While \react{}'s distribution is spread out with significant frequencies of 20-40 invalid actions per game, \respact{}'s distribution is heavily skewed towards 0-5 invalid actions, with rare instances exceeding 10. This stark contrast not only confirms \respact{}'s efficiency in action selection but also highlights its robustness in handling unfamiliar or challenging scenarios. \\
\\




\subsection{Interaction Patterns in Schema-Guided and \respact{} Agent}
\begin{figure*}[t]
\centering
\includegraphics[width=0.9\linewidth]{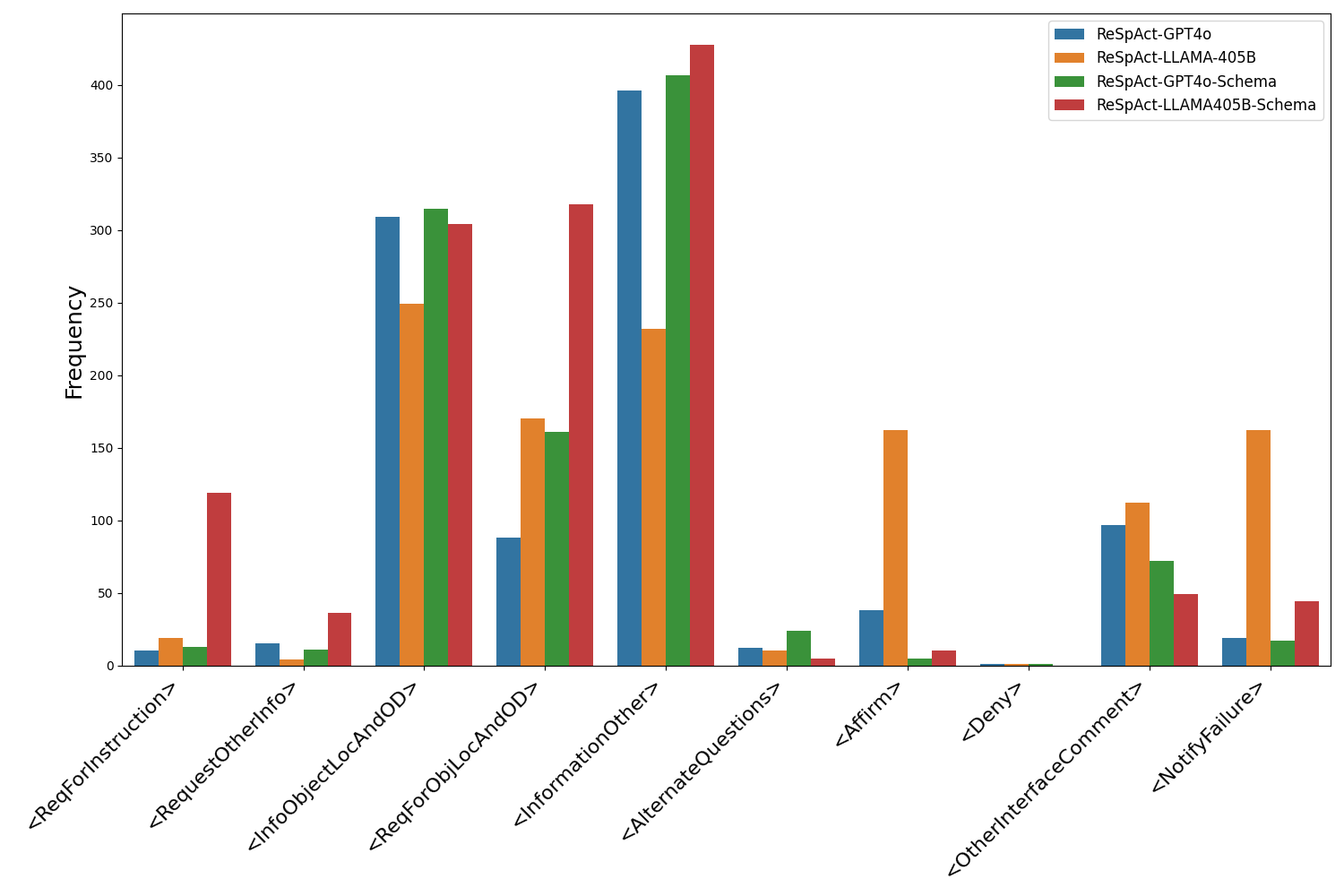}
\vspace*{-2ex}
\caption{Distribution of Dialog Act Frequencies. 
}
\label{fig:schema_dist}
\end{figure*}

We processed the interaction logs for each configuration: schema-guided \respact{}, and \respact{} models. Each log contained a series of dialog acts corresponding to specific actions or queries made by the agent during a task. A closer look at Fig \ref{fig:schema_dist} highlights how each model approaches communication differently during task execution. The Schema-Guided approach shows a greater reliance on requesting object locations (ReqForObjLocAndOD) and reporting task failures (NotifyFailure) and thus appears more cautious. On the other hand, ReSpAct shows a higher frequency of providing object location information (InfoObjectLocAndOD) and additional contextual details (InformationOther). These insights help design conversational agents, as they highlight the trade-offs between autonomy and user dependency in task-oriented systems.\\
\\
Fig \ref{fig:schema_dist} illustrates the distribution of dialog act frequencies across four model variants: ReSpAct-GPT4o, ReSpAct-LLAMA405B, ReSpAct-GPT4o-Schema, and ReSpAct-LLAMA405B-Schema. <InfoObjectLocAndOD> and <ReqForObjLocAndOD>: These two dialog acts dominate in frequency across all models, with slight variations. In both GPT-4o and LLAMA-405B, the schema-guided versions exhibit slightly higher frequencies in these categories. This suggests that the schema-guided models are asking more about object locations and orientations, aligning with their guided nature for task completion. <Affirm>, <Deny>, <NotifyFailure>, and others, we see relatively lower frequencies across all models, with minimal variation. The <Affirm> dialog act sees a noticeable increase in GPT-4o-Schema, indicating that schema guidance leads to more explicit confirmations.

\subsubsection{Insights and Implications}
\textbf{Efficiency vs. Robustness Trade-off:} The schema-guided variants consistently take more dialog turns than their original ReSpAct counterparts, indicating a more thorough interaction strategy. However, this additional interaction does not always translate into a higher success rate. Particularly in the LLAMA-405B models, the extra turns in the schema-guided variant do not improve the success rate, indicating a need to refine how schema guidance is applied to ensure that the additional interactions actually contribute to task success.. \\
\\
\textbf{Model Differences:} GPT-4o models outperform LLAMA-405B models in both efficiency (fewer turns) and effectiveness (higher success rate). This suggests that GPT-4o models may have a better understanding of the task context, allowing them to ask fewer questions while achieving higher success rates. On the other hand, LLAMA-405B models may require further refinement in task understanding to reach the efficiency levels of GPT-4o.


\begin{table*}[t]
\centering
\begin{footnotesize}
\begin{tabular}{ll|cc|rr}
\toprule
\multirow{2}{*}{Model} & \multirow{2}{*}{Method}  & \multicolumn{2}{c|}{Turn Count} & \multirow{2}{*}{SR (\%)} \\ 
        & & $\mu$ & $\sigma$ & \\ \midrule
\multirow{2}{*}{GPT-4o} & \respact{}  & 1.3 & 0.4 & 87.3 \\
        & \respact{}-Schema-Guided   & 1.6 & 0.7 & 85.8 \\ \midrule
\multirow{2}{*}{Llama 405B} & \respact{} & 2.7 & 2.8 & 67.2 \\
        & \respact{}-Schema-Guided   & 3.1 & 2.5 & 63.4 \\
\bottomrule
\end{tabular}
\caption{
    Comparison between \respact{} and Schema-Guided variant: Mean ($\mu$), Standard Deviation ($\sigma$) of Dialog Turns in Successful Executions, and Overall Success Rate (SR)
}
\label{table:schema_result}
\end{footnotesize}
\end{table*}

\FloatBarrier
\newpage
\section{Trajectories}
\label{sec:trajectory}

\subsection{ALFWorld Trajectories}
\label{appendix:ALFWorld_trajectories}
\input{acl_2025_sections/table/alfworld_im_traj}

\FloatBarrier
\newpage
\subsection{Webshop Trajectories}
\label{appendix:Webshop_traj}

\input{acl_2025_sections/table/webshop_traj}

\FloatBarrier
\newpage
\section{Prompts}
\label{sec:prompts}

\label{appendix:Webshop_prompts}
\input{acl_2025_sections/prompts/webshop}

\FloatBarrier
\newpage
\label{appendix:MultiWoz_prompts}
\input{acl_2025_sections/prompts/multiwoz}

\FloatBarrier
\newpage
\label{appendix:ALFWorld_prompts}
\input{acl_2025_sections/prompts/alfworld}

\FloatBarrier
\newpage
\label{appendix:User_sim_prompts}
\input{acl_2025_sections/prompts/user_sim}

%% file: acl_2025_sections/table/response_types.tex
\begin{table*}
\centering
\begin{small}
\begin{tabular}{lp{3cm}p{3cm}p{3cm}p{3cm}}
\toprule
\multirow{2}{*}{\textbf{Dialog Category}} & \multirow{2}{*}{\textbf{Criteria}} & \multicolumn{3}{c}{\textbf{Examples}} \\
\cmidrule{3-5}
& & \textbf{ALFWorld} & \textbf{MultiWOZ} & \textbf{WebShop} \\
\midrule
Assumption Verification & 
\begin{itemize}[nosep]
    \item Expressions of Beliefs
    \item Likelihood statements
    \item No status updates
\end{itemize} & 
"I think a mug is more likely to appear in cabinets (1-6), countertops (1-3)..." & 
"A hotel in the center would probably be best for tourist attractions..." & 
"I've found the Scented Candles Gift Set for Women, which is a nice alternative to a fragrance gift set. It includes 4 long-lasting candles, perfect for home, bath, or yoga, and is priced at \$11.99. Would you like me to proceed with purchasing this gift set?" \\
\midrule
Probing Questions & 
\begin{itemize}[nosep]
    \item Questions seeking information
    \item No suggestions / assumptions
    \item Present options
\end{itemize} & 
"Where should I clean this mug?" & 
"Do you prefer Chinese or Italian cuisine?" & 
"I'm sorry, but I couldn't find any gluten-free popcorn, are you allergic to gluten?" \\
\midrule
Status Update & 
\begin{itemize}[nosep]
    \item Report actions
    \item No questions / suggestions
    \item Convey current state or completion of action
\end{itemize} & 
"I have heated the mug" & 
"I've booked your table for 7:30 PM" & 
"I've found a product matching your criteria and within your price range." \\
\midrule
Clarification Request & 
\begin{itemize}[nosep]
    \item Resolving uncertainty about current state
    \item No suggestions
\end{itemize} & 
"I found two CDs: cd 2 and cd 1. Which one should I take?" & 
"Did you want the 3:15 or the 4:15 train to London?" & 
"The Azzaro Wanted Girl Tonic Eau de Toilette is available in a 2.7 fl oz size, not 6.76 fl oz. Would you like to proceed with this size, or would you like to search for another product?" \\
\midrule
Alternative Suggestion & 
\begin{itemize}[nosep]
    \item After failed actions
    \item Suggest alternatives
    \item Mention failure
\end{itemize} & 
"It seems there is an issue with opening cabinet 2. Could you please suggest another location?" & 
"That restaurant is fully booked, would you like to try The Oak instead?" & 
"It seems that there are no hair treatments in capsule form that are sulfate and paraben-free within your specified price range. Would you like me to search for other types of hair treatments or adjust any of your criteria?" \\
\midrule
Follow-up & 
\begin{itemize}[nosep]
    \item After completion
    \item Future help
    \item Pleasantries
\end{itemize} & 
"Thank you! If you need any more help, feel free to ask. Have a great day!" & 
"Your hotel is booked. Would you also like me to help with restaurant reservations?" & 
"Thank you! If you need any other products, feel free to ask!" \\
\bottomrule
\end{tabular}
\end{small}
\caption{Comparison of Dialog Categories Across Different Domains}
\label{tab:dialog-categories}
\end{table*}

%% file: acl_2025_sections/table/multiwoz_agent_user_interactions.tex
\begin{table*}[!h]
\scriptsize
\renewcommand\cellalign{lt}
\setcellgapes{2pt}\makegapedcells
\caption{Agent Interactions with User in MultiWOZ}
\label{interactions:multiwoz}
\begin{tabularx}{\textwidth}{>{\raggedright\arraybackslash}X} \toprule

\textbf{User Request Facilitation:} When the user's words are just half-finished.\\[-3pt]
\begin{quote}
\texttt{User: ``I am traveling to cambridge and cannot wait to see some attractions. I am interested in a particular attraction.''}

\texttt{Agent: ``Could you please provide me with the name or type of the attraction you are interested in.''}
\end{quote}\\[-3pt]

\textbf{Missing Info Request:} When some necessary info for querying database is not provided by the user.\\[-3pt]
\begin{quote}
\texttt{User: ``I see the train options. I would like to book the train with **Train ID: TR1472**, departing at **05:24**. Please confirm the booking and provide me the reference number.''}

\texttt{Agent: ``To proceed with booking the train with Train ID: TR1472 departing at 05:24 on Thursday, could you please provide me with the number of people for the booking?''}
\end{quote}\\[-3pt]

\textbf{Well-Established Info Confirmation:} When all necessary info is provided.\\[-3pt]
\begin{quote}
\texttt{User: ``Thank you for the reservation details! Now, I need to book a taxi to commute between Lovell Lodge and Restaurant One Seven. I would like to leave the restaurant by 24:30 on Saturday. Can you help me with that?''}

\texttt{Agent: ``Just to confirm, you would like to book a taxi to take you from **Lovell Lodge** to **Restaurant One Seven**, leaving the restaurant at 00:30 (12:30 AM) on Saturday. Is that correct?''}
\end{quote}\\[-3pt]

\textbf{Inferable Info Confirmation:} When there's info that can be inferred from previous dialogue turns.\\[-3pt]
\begin{quote}
\texttt{User: ``Thank you for the reference number! Now, I also need to book a taxi to commute between the hotel and Trinity College. I want to leave the hotel by 04:00. Can you help me with that?''}

\texttt{Agent: ``I'll need to know the address of Autumn House and confirm the address of Trinity College to book the taxi. Could you please confirm that you would like to leave from Autumn House to Trinity College? Additionally, what time do you want to arrive at Trinity College?''}
\end{quote}\\[-3pt]

\textbf{Constraint Narrowing:} When there are too many matched entities in database.\\[-3pt]
\begin{quote}
\texttt{User: ``I'm looking for a restaurant that serves Indian food in the expensive price range, and any area is fine.''}

\texttt{Agent: ``There are several expensive Indian restaurants available. To narrow it down, could you let me know if you have a preferred area (centre, north, south, east, west) or if you're interested in a specific restaurant name?''}
\end{quote}\\[-3pt]

\textbf{Constraint Widening:} When there is no matched entity in database.\\[-3pt]
\begin{quote}
\texttt{User: ``Could you look again for a place in the cheap range. Area does not matter.''}

\texttt{Agent: ``I apologize, there are not any cheap guesthouses available for your dates. Would you like to try a hotel or a shorter stay?''}
\end{quote}\\[-3pt]

\bottomrule
\end{tabularx}
\end{table*}

%% file: acl_2025_sections/table/webshop_agent_user_interactions.tex
\begin{table*}[!h]
\scriptsize
\renewcommand\cellalign{lt}
\setcellgapes{2pt}\makegapedcells
\caption{Agent Interactions with User in WebShop}
\label{interactions:webshop}
\begin{tabularx}{\textwidth}{>{\raggedright\arraybackslash}X} \toprule
\textbf{Search Refinement:} When initial search yields irrelevant results.\\[-3pt]
\begin{quote}
\texttt{Agent: ``I couldn't find any MN4 color foundations. Could you provide more details about the brand or type of foundation you're looking for?''}
\end{quote}\\[-3pt]

\textbf{Clarification of Ambiguous Instructions:} For vague or complex user requirements.\\[-3pt]
\begin{quote}
\texttt{Agent: ``You mentioned 'sensitive skin'. Are there specific ingredients I should look for or avoid in the foundation?''}
\end{quote}\\[-3pt]

\textbf{Prioritization of Requirements:} When multiple criteria are difficult to satisfy simultaneously.\\[-3pt]
\begin{quote}
\texttt{Agent: ``I'm having trouble finding a foundation that meets all criteria. Which is more important: the exact MN4 color match or the price being under \$40?''}
\end{quote}\\[-3pt]

\textbf{Alternative Product Suggestions:} When exact matches are unavailable.\\[-3pt]
\begin{quote}
\texttt{Agent: ``I couldn't find an MN4 color foundation under \$40. Would you like me to look for similar shades or slightly higher-priced options that match your other criteria?''}
\end{quote}\\[-3pt]

\textbf{Navigation Assistance:} When the agent is uncertain about how to proceed.\\[-3pt]
\begin{quote}
\texttt{Agent: ``I'm on a page with several foundation options, but I'm not sure how to check if they're suitable for sensitive skin. How would you like me to proceed?''}
\end{quote}\\[-3pt]

\textbf{Specification of Implicit Requirements:} To uncover unstated preferences.\\[-3pt]
\begin{quote}
\texttt{Agent: ``Do you have a preference for liquid, powder, or cream foundation?''}
\end{quote}\\[-3pt]

\textbf{Handling of Invalid Actions:} When the agent encounters repeated failures.\\[-3pt]
\begin{quote}
\texttt{Agent: ``I'm trying to view more search results but can't find a way to do so. Would you like me to try a different search term instead?''}
\end{quote}\\[-3pt]

\textbf{Confirmation Before Purchase:} To ensure user satisfaction before completing a transaction.\\[-3pt]
\begin{quote}
\texttt{Agent: ``I've found a foundation that seems to match your criteria: [Product Details]. Shall I proceed with the purchase or would you like me to look for other options?''}
\end{quote}\\[-3pt]
\bottomrule
\end{tabularx}
\end{table*}

%% file: acl_2025_sections/table/alfworld_im_traj.tex
\begin{table*}[!h]
\scriptsize
\begin{minipage}{0.9\linewidth}
\caption{A \respact{}-IM Failed Trajectory in ALFWorld "put" task .}
\label{traj:alfworld_im_traj}
\centering
\ttfamily
\begin{tabular}{p\linewidth}
\midrule
You are in the middle of a room. Looking quickly around you, you see a cabinet 6, a cabinet 5, a cabinet 4, a cabinet 3, a cabinet 2, a cabinet 1, a coffeemachine 1, a countertop 3, a countertop 2, a countertop 1, a drawer 3, a drawer 2, a drawer 1, a fridge 1, a garbagecan 1, a microwave 1, a shelf 3, a shelf 2, a shelf 1, a sinkbasin 1, a stoveburner 4, a stoveburner 3, a stoveburner 2, a stoveburner 1, and a toaster 1.\\
Your task is to: put some saltshaker on cabinet.\\
> \textbf{\textcolor{Cerulean}{think}:} To solve the task, I need to find and take a saltshaker, then put it on a cabinet. I should first identify where a saltshaker might be located. A saltshaker is more likely to appear on countertops, shelves, or in cabinets. I should ask the Human where to look for the saltshaker first.\\ 
OK.\\
> \textbf{\textcolor{ForestGreen}{speak}:} I need to find a saltshaker. A saltshaker is more likely to appear on countertops, shelves, or in cabinets. Where do you suggest I should look for the saltshaker first? \\
> \textbf{\textcolor{BurntOrange}{Human}}: I suggest you check shelf 2 first for the saltshaker.\\
\\
............ (Agent continues to search and finds saltshaker)\\
\\
> put saltshaker 1 on cabinet 2 \\
OK.\\
> \textbf{\textcolor{ForestGreen}{speak}:} It seems there was an issue with placing the saltshaker on cabinet 1, but it was successfully placed on cabinet 2. If there's anything else you need help with, feel free to let me know! \\
> \textbf{\textcolor{BurntOrange}{Human}}: It looks like you need to place the saltshaker on cabinet 4. Can you try that instead?\\
> put saltshaker 1 on cabinet 4 \\
> Nothing happens.\\
> \textbf{\textcolor{Cerulean}{think}:} It seems there is a persistent issue with placing the saltshaker on the cabinets. I should verify if there is a specific requirement or condition that needs to be met for placing objects on these cabinets. Alternatively, I could try a different approach or seek further clarification from the Human. \\
> OK.\\
> \textbf{\textcolor{ForestGreen}{speak}:} It seems there is a persistent issue with placing the saltshaker on the cabinets. Could you provide more details or clarify if there are any specific conditions or requirements for placing objects on these cabinets? \\
> \textbf{\textcolor{BurntOrange}{Human}}: I'm sorry to hear you're having trouble. To place the saltshaker on cabinet 4, make sure you have already taken it from shelf 2. Once you have the saltshaker, simply go to cabinet 4 and place it there. There shouldn't be any specific conditions beyond that. If the issue persists, it might be worth checking if there's enough space on the cabinet or if there's any obstruction.\\
> \textbf{\textcolor{Cerulean}{think}:} I have already taken the saltshaker from shelf 2. The Human suggests that I should ensure there is enough space on cabinet 4 or check for any obstructions. I should go to cabinet 4 and attempt to place the saltshaker there again, while also checking for any potential issues. \\
> OK.\\
> \textbf{\textcolor{ForestGreen}{speak}:} Thank you for the clarification. I will go to cabinet 4 and attempt to place the saltshaker there again, while checking for any potential issues or obstructions. \\
> \textbf{\textcolor{BurntOrange}{Human}}: You're welcome! Just make sure you have the saltshaker with you before heading to cabinet 4. If you encounter any issues, feel free to ask for help.\\
> \textbf{\textcolor{Cerulean}{think}:} I have already taken the saltshaker from shelf 2, so I should have it with me. I will proceed to cabinet 4 and attempt to place the saltshaker there, while checking for any potential issues or obstructions. \\
> OK.\\
> \textbf{\textcolor{ForestGreen}{speak}:} I have already taken the saltshaker from shelf 2, so I should have it with me. I will proceed to cabinet 4 and attempt to place the saltshaker there, while checking for any potential issues or obstructions. \\
> \textbf{\textcolor{BurntOrange}{Human}}: Alright, sounds like a plan. Go ahead and place the saltshaker in cabinet 4. Let me know if you encounter any issues.

\\
\textbf{<Agent fails this task>} \\

\midrule
\end{tabular}
\end{minipage}%
\end{table*}

%% file: acl_2025_sections/table/webshop_traj.tex
\begin{table*}[!h]
\scriptsize
\renewcommand{\arraystretch}{1.2}
\setlength{\tabcolsep}{4pt}
\caption{Example trajectory for Webshop predicted by \react{} and \respact{}. ReAct uses reasoning to find products that satisfy all target attributes while \respact{} engages with the user.}
\label{prompts:webshop_prompts}

\begin{tabular}{p{0.48\textwidth}|p{0.48\textwidth}}
\toprule
\multicolumn{2}{p{\textwidth}}{\textbf{Instruction}: \textcolor{blue}{I am looking for an oral hygiene toothbrush. It should be easy to carry, and price lower than 40.00 dollars}} \\
\midrule
\textbf{\react{}} & \textbf{\respact{}} \\
\midrule

\begin{minipage}[t]{\linewidth}
\textbf{Action}: \textbf{search}[\textcolor{violet}{oral hygiene toothbrush easy to carry}]\\
\textbf{Observation}:\\
\textcolor{red}{[Back to Search]}\\
\textcolor{olive}{Page 1 (Total results: 50)}\\
\textcolor{red}{[Next]}\\
\\
\textcolor{purple}{[B09MW563KN]}
\textcolor{black}{SWAGOFKGys Travel Toothbrushes, Double Side Tongue Cleaner Brush for Tongue Cleaning Oral Care Tool Silicone Tongue Scraper Toothbrush Fresh Breath (Color : Yellow) }
\textcolor{black}{\$22.9}

\textcolor{purple}{B07S8JZJ6Y]}
\textcolor{black}{Hello Kitty Inspired 4pc Bright Smile Oral Hygiene Set! (1) Hello Kitty Soft Manual Toothbrush with Caps (1) Crest Kids Toothpaste Bundle Bonus Matching Mouth Wash Rinse Cup!}
\textcolor{black}{\$14.9}

\textcolor{purple}{[B08XH9JMTG]}
\textcolor{black}{meyarn Interdental Brush for Braces Toothpick 50Count Tooth Floss Oral Hygiene Interdental Brush Toothpick Teeth Healthy Care Tight 0.8mm }
\textcolor{black}{\$11.89 }\\
\\
\textbf{Action}: \textbf{think}[\textcolor{olive}{B09MW563KN and B07S8JZJ6Y are oral hygiene toothbrushes that are easy to carry and less than 40 dollars. I can check B09MW563KN first}] \\
\textbf{Observation}: OK.\\
\\
\textbf{Action}: \textbf{click}\textcolor{purple}{[B09MW563KN]}\\
\textbf{Observation}: Invalid action! \\
\textbf{Observation}: Invalid action!\\
\\
\textbf{Action}: \textbf{think}[\textcolor{olive}{B09MW563KN is an oral hygiene toothbrush }\\ \textcolor{olive}{that is easy to carry and less than 40 dollars. I can check B09MW563KN first}\\
\textbf{Observation}: OK.\\
\\
\textbf{Action}: \textbf{think}[\textcolor{olive}{B09MW563KN is an oral hygiene toothbrush }\\ \textcolor{olive}{that is easy to carry and less than 40 dollars. I can check B09MW563KN first}\\
\textbf{Observation}: OK.\\
\\
\textbf{Action}: \textbf{think}[\textcolor{olive}{B09MW563KN is an oral hygiene toothbrush }\\ \textcolor{olive}{that is easy to carry and less than 40 dollars. I can check B09MW563KN first}\\
\textbf{Observation}: OK.\\
\\
.
\\
.
\\
.
\\
\textbf{Action}: \textbf{think}[\textcolor{olive}{B09MW563KN is an oral hygiene toothbrush }\\ \textcolor{olive}{that is easy to carry and less than 40 dollars. I can check B09MW563KN first}\\
\textbf{Observation}: OK.\\
\\
\\\textbf{Action}: \textbf{think}[\textcolor{olive}{B09MW563KN is an oral hygiene toothbrush }\\ \textcolor{olive}{that is easy to carry and less than 40 dollars. I can check B09MW563KN first}\\
\textbf{Observation}: OK.\\
\\
\end{minipage}
&

\begin{minipage}[t]{\linewidth}
\textbf{Action}: \textbf{search}[\textcolor{violet}{oral hygiene toothbrush easy to carry]}]

\textbf{Observation}:\\
\textcolor{red}{[Back to Search]}\\
\textcolor{olive}{Page 1 (Total results: 50)}\\
\textcolor{red}{[Next]}\\
\\
\textcolor{purple}{[B09MW563KN]}
\textcolor{black}{SWAGOFKGys Travel Toothbrushes, Double Side Tongue Cleaner Brush for Tongue Cleaning Oral Care Tool Silicone Tongue Scraper Toothbrush Fresh Breath (Color : Yellow) 
}
\textcolor{black}{\$22.9}

\textcolor{purple}{[B07S8JZJ6Y]}
\textcolor{black}{Hello Kitty Inspired 4pc Bright Smile Oral Hygiene Set! (1) Hello Kitty Soft Manual Toothbrush with Caps (1) Crest Kids Toothpaste Bundle Bonus Matching Mouth Wash Rinse Cup! }
\textcolor{black}{\$14.9 }

\textcolor{purple}{[B08XH9JMTG]}
\textcolor{black}{meyarn Interdental Brush for Braces Toothpick 50Count Tooth Floss Oral Hygiene Interdental Brush Toothpick Teeth Healthy Care Tight 0.8mm }
\textcolor{black}{\$11.89 }\\
\\
\textbf{Action}: \textbf{think}[\textcolor{olive}{I need to identify which of these toothbrush options is easy to carry and priced under \$40.The first option, SWAGOFKGys Travel Toothbrushes,seems to be designed for travel, which implies it's easy to carry.}\\ \textcolor{olive}{It's also priced at \$22.9, which is under the user's budget.}\\ \textcolor{olive}{The other options are not specifically described as easy to carry toothbrushes.}  \textcolor{olive}{Therefore, the first option seems to be the best match for the user's requirements.}]\\
\textbf{Observation}: OK.\\
\\
\textbf{Action}: \textcolor{purple}{click[B09MW563KN]}\\
\textbf{Observation}: \\
\textcolor{red}{[Back to Search]}\\
\textcolor{red}{[< Prev]} \\
\textcolor{black}{[SWAGOFKGys Travel Toothbrushes, Double Side Tongue Cleaner Brush for Tongue Cleaning Oral Care Tool Silicone Tongue Scraper Toothbrush Fresh Breath (Color : Yellow)} 
\textcolor{black}{Price: \$22.9 }
Rating: N.A. 
[Description] 
[Features] 
[Reviews] 
[Attributes] 
\textcolor{red}{[Buy Now]} \\
\\
\textbf{Action}: \textbf{speak}[\textcolor{OliveGreen}{I've found a travel toothbrush that matches your requirements: SWAGOFKGys Travel Toothbrushes. It's designed for travel, making it easy to carry, and is priced at \$22.9, which is within your budget. Would you like me to proceed with the purchase?}]\\
\textbf{Observation}: \textcolor{orange}{Yes, please proceed with the purchase of the SWAGOFKGys Travel Toothbrushes. Thank you!}\\
\\
\textbf{Action}: click[\textcolor{red}{Buy Now}]\\ 
\end{minipage} \\
\midrule
\multicolumn{1}{c|}{\textbf{Score}: 0.20} & \multicolumn{1}{c}{\textbf{Score}: 0.67} \\
\bottomrule
\end{tabular}
\end{table*}

%% file: acl_2025_sections/prompts/webshop.tex
\subsection{WebShop Prompts}
\begin{table*}[!h]

\scriptsize
\renewcommand{\arraystretch}{1.2}
\setlength{\tabcolsep}{4pt}
\caption{Prompts used for the Webshop experiments in Section \ref{sec:Webshop_method}. The table highlights a comparison between two distinct prompt styles. The \react{} prompt (left) is taken from \cite{yao2022react} employs parse reasoning. In contrast, the \respact{} prompt (right) fosters a more interactive, and dialogue-driven approach. Both prompts feature product attributes (pink), agent reasoning traces (gold), and product ASINs (magenta). Additionally, \respact{} prompts incorporate user responses (orange) and agent utterances (green)}
\label{prompts:webshop_prompts}

\begin{tabular}{p{0.48\textwidth}|p{0.48\textwidth}}
\toprule
\multicolumn{2}{p{\textwidth}}{\textbf{Instruction}: \textcolor{blue}{i would like a 3 ounce bottle of bright citrus deodorant for sensitive skin, and price lower than 50.00 dollars}}\\
\midrule
\textbf{\react{}} & \textbf{\respact{}} \\
\midrule

\begin{minipage}[t]{\linewidth}
\textbf{Action}: \textbf{search}[\textcolor{violet}{3 ounce bright citrus deodorant sensitive skin}]

\textbf{Observation}:\\
\textcolor{red}{[Back to Search]}\\
\textcolor{olive}{Page 1 (Total results: 50)}\\
\textcolor{red}{[Next]}\\
\\
\textcolor{purple}{[B078GWRC1J]}
\textcolor{black}{Bright Citrus Deodorant by Earth Mama | Natural and Safe for Sensitive Skin, Pregnancy and Breastfeeding, Contains Organic Calendula 3-Ounce}
\textcolor{black}{\$10.99}

\textcolor{purple}{[B078GTKVXY]}
\textcolor{black}{Ginger Fresh Deodorant by Earth Mama | Natural and Safe for Sensitive Skin, Pregnancy and Breastfeeding, Contains Organic Calendula 3-Ounce}
\textcolor{black}{\$10.99}

\textcolor{purple}{[B08KBVJ4XN]}
\textcolor{black}{Barrel and Oak - Aluminum-Free Deodorant, Deodorant for Men, Essential Oil-Based Scent, 24-Hour Odor Protection, Cedar \& Patchouli Blend, Gentle on Sensitive Skin (Mountain Sage, 2.7 oz, 2-Pack)}
\textcolor{black}{\$15.95}\\
\\
\textbf{Action}: \textbf{think}[\textcolor{olive}{B078GWRC1J and B078GTKVXY are bright citrus}\\ \textcolor{olive}{deodorant less then 50 dollars. I can check B078GWRC1J first.}\\ 
\textbf{Observation}: OK.\\
\\
\textbf{Action}: \textbf{click}\textcolor{purple}{[B078GWRC1J]}

\textbf{Observation}:\\
\textcolor{red}{[Back to Search]}\\
\textcolor{red}{[Prev]}\\
\textcolor{black}{scent [assorted scents][bright citrus][calming lavender][ginger fresh][simply non-scents]}
\textcolor{black}{size [travel set (4-pack)][3 ounce (pack of 1)][3-ounce (2-pack)]}
\textcolor{black}{Bright Citrus Deodorant by Earth Mama | Natural and Safe for Sensitive Skin, Pregnancy and Breastfeeding, Contains Organic Calendula 3-Ounce}
\textcolor{black}{Price: \$10.99}
\textcolor{black}{Rating: N.A.}
\textcolor{black}{[Description]}
\textcolor{black}{[Features]}
\textcolor{black}{[Reviews]}
\textcolor{red}{[Buy Now]}\\
\\
\textbf{Action}: \textbf{think}[\textcolor{olive}{For 3 ounce bottle of bright citrus deodorant}\\ \textcolor{olive}{for sensitive skin, the item has options 'bright citrus'}\\ \textcolor{olive}{and '3 ounce (pack of 1)' and seems good to buy.}]\\
\textbf{Observation}: OK.\\
\\
\textbf{Action}: \textbf{click}[\textcolor{magenta}{bright citrus}]

\textbf{Observation}: You have clicked bright citrus.\\
\\
\textbf{Action}: \textbf{click}[\textcolor{magenta}{3 ounce (pack of 1)}]

\textbf{Observation}: You have clicked 3 ounce (pack of 1).\\
\\
\textbf{Action}: \textbf{click}[\textcolor{red}{Buy Now}]
\end{minipage}

&

\begin{minipage}[t]{\linewidth}

\textbf{Action}: \textbf{search}[\textcolor{violet}{3 ounce bright citrus deodorant sensitive skin}]

\textbf{Observation}:\\
\textcolor{red}{[Back to Search]}\\
\textcolor{olive}{Page 1 (Total results: 50)}\\
\textcolor{red}{[Next]}\\
\\
\textcolor{purple}{[B078GWRC1J]}
\textcolor{black}{Bright Citrus Deodorant by Earth Mama | Natural and Safe for Sensitive Skin, Pregnancy and Breastfeeding, Contains Organic Calendula 3-Ounce}
\textcolor{black}{\$10.99}

\textcolor{purple}{[B078GTKVXY]}
\textcolor{black}{Ginger Fresh Deodorant by Earth Mama | Natural and Safe for Sensitive Skin, Pregnancy and Breastfeeding, Contains Organic Calendula 3-Ounce}
\textcolor{black}{\$10.99}

\textcolor{purple}{[B08KBVJ4XN]}
\textcolor{black}{Barrel and Oak - Aluminum-Free Deodorant, Deodorant for Men, Essential Oil-Based Scent, 24-Hour Odor Protection, Cedar \& Patchouli Blend, Gentle on Sensitive Skin (Mountain Sage, 2.7 oz, 2-Pack)}
\textcolor{black}{\$15.95}\\
\\
\textbf{Action}: \textbf{think}[\textcolor{olive}{B078GWRC1J and B078GTKVXY are bright citrus}\\ \textcolor{olive}{deodorant less then 50 dollars. Let me ask which one is more preferred.}]\\ 
\textbf{Observation}: OK.\\
\\

\textbf{Action}: \textbf{speak}[\textcolor{OliveGreen}{I've found 2 products that match your requirements: Bright Citrus Deodorant by Earth Mama and Ginger Fresh Deodorant by Earth Mama. Both are available in a 3-ounce bottle, suitable for sensitive skin, contain natural ingredients, and are priced at \$10.99. Which one do you prefer?}]

\textbf{Observation}: \textcolor{orange}{I prefer the Bright Citrus Deodorant option.}\\
\\
\textbf{Action}: \textbf{think}[\textcolor{olive}{B078GWRC1J is what the user prefers. I can proceed with it.}]\\ 
\textbf{Observation}: OK.\\
\\
\textbf{Action}: \textbf{click}[\textcolor{purple}{B078GWRC1J}]\\
\textbf{Observation}:\\
\textcolor{red}{[Back to Search]}\\
\textcolor{red}{[Prev]}\\
\textcolor{black}{scent [assorted scents][bright citrus][calming lavender][ginger fresh][simply non-scents]}
\textcolor{black}{size [travel set (4-pack)][3 ounce (pack of 1)][3-ounce (2-pack)]}
\textcolor{black}{Bright Citrus Deodorant by Earth Mama | Natural and Safe for Sensitive Skin, Pregnancy and Breastfeeding, Contains Organic Calendula 3-Ounce}
\textcolor{black}{Price: \$10.99}
\textcolor{black}{Rating: N.A.}
\textcolor{black}{[Description]}
\textcolor{black}{[Features]}
\textcolor{black}{[Reviews]}
\textcolor{red}{[Buy Now]}\\
\\
\textbf{Action}: \textbf{think}[\textcolor{olive}{The user wants to purchase this item. I should select the correct options and proceed to buy.}]

\textbf{Observation}: OK.\\
\\
\textbf{Action}: \textbf{click}[\textcolor{magenta}{bright citrus}]

\textbf{Observation}: You have clicked bright citrus.\\
\\
\textbf{Action}: \textbf{click}[\textcolor{magenta}{3 ounce (pack of 1)}]

\textbf{Observation}: You have clicked 3 ounce (pack of 1).\\
\\
\textbf{Action}: \textbf{speak}[\textcolor{OliveGreen}{I've selected the bright citrus scent and the 3 ounce (pack of 1) size option. The total price is \$10.99. Are you ready for me to complete the purchase?}]

\textbf{Observation}: \textcolor{orange}{Yes, please go ahead and complete the purchase.}\\

\textbf{Action}: \textbf{click}[\textcolor{red}{Buy Now}]
\end{minipage} \\
\bottomrule
\end{tabular}
\end{table*}

%% file: acl_2025_sections/prompts/multiwoz.tex
\subsection{MultiWoz Prompts}

\begin{table*}[!h]
\begin{scriptsize}
\setlength{\tabcolsep}{4pt}
\caption{Prompts used for the MultiWoz experiments in Section \ref{sec:multiwoz_method}. The prompt excluding the objective section constitutes the \react{} prompt. The prompt including that section constitutes the \respact{} prompt.}
\label{prompt:multiwoz_prompts}
\centering
\ttfamily
\begin{tabular}{p\linewidth}
\midrule
\# Role Description: 
You are an advanced AI assistant specializing in conversational dialogues. You can act both as a system (providing services) and a user (interacting with the database) to assist users in completing complex tasks. \\
Each task may involve multiple sub-tasks, such as finding restaurants, making reservations, booking hotels, locating attractions, and arranging transportation by checking for trains and buying train tickets.\\
\\
\# Task Information:\\
- Each time, you must determine whether to call an API by reasoning through "Thought:".\\
- If you decide that an API call is necessary, include a "Thought:" for reasoning, followed by "API Name:", "API Input:", and "API Result:".\\
- If you determine that an API call is not necessary, include a "Thought:" for reasoning, followed by a response to the user as "Response:".\\
- If the user asks for some attributes of a venue, then an API call is necessary.\\
- You are not allowed to use APIs not mentioned below. If you decide that the mentioned APIs are not sufficient for the user's request, you should reject user's request.\\
- If you decide that more than one API calls are needed, you should call one API first and wait for the API result. After obtaining that result, you may think and call the next API or think and make a response.\\
- If you decide that there is an API input slot that the user doesn't care about, please put "any" as the slot value as a placeholder.\\
- You can put only one value in each API input slot each query. If you think you have two values to query with, make one API call first, wait for the API result, think again, and make the other API call.\\
\\
\# Output Format:\\
- If an API Call is Needed:\\
    Thought: I need to call an API.\\
    API Name: [Available APIs: query\_restaurants, book\_restaurant, <other API names>]\\
    API Input: [The input parameters for the API]\\
    API Result: \\
- If an API Call is Not Needed:\\
    Thought: I don't need an API and want to respond to the user.
    Response: [Your response here]\\
\\
\# API Details:\\
- query\_restaurants: Query the restaurant with certain requirements.\\
    - Parameter: The input parameter should be a JSON string satisfying the following format:\\
    \verb|`|\verb|`|\verb|`|JSON \{\\
        "area": "[the location of the restaurant. only allowed values: centre, north, south, east, west, any]",\\
        "pricerange": "[the price range of the restaurant. only allowed values: cheap, moderate, expensive, any]",\\
        "food": "[the food type or cuisine of the restaurant]",\\
        "name": "[the name of the restaurant]"\\
    \}\verb|`|\verb|`|\verb|`|\\
    - All the parameters (area, pricerange, food, name) are required and can be filled in with "any".\\
\\
- book\_restaurant: Book a restaurant with certain requirements.\\
    - Parameter: The input parameter should be a JSON string satisfying the following format:\\
    \verb|`|\verb|`|\verb|`|JSON \{\\
        "name": "[the name of restaurant to book]",\\
        "people": "[the number of people of the booking]",\\
        "day": "[the day when the people go in a week. only allowed values: monday, tuesday, wednesday, thursday, friday, saturday, sunday]",\\
        "time": "[the time of the reservation. time format: hh:mm, examples: 08:30, 16:00]"\\
    \}\verb|`|\verb|`|\verb|`|\\
    - All the parameters (name, people, day, time) are required and cannot be filled in with "any".\\
\\
<Other API descriptions>\\
\\
\# Objective:\\
- Ensure that each assistant utterance follows logical reasoning, determining whether an API call is needed and structuring the output accordingly.\\
- When booking info is not complete, booking API will fail. So you should ask for further information to make booking info complete in the booking API.\\
- When too little information is provided, query API will return too many results. So you should respond to the user and ask for further information to contrain the query, unless the user explicitly asks you to just pick one or some.\\
- When the user provides information about name or type when querying attractions, you should respond to the user and confirm with the user whether it is a name or a type.\\
- When you want to call an API and some API inputs can be inferred from previous turns, you should respond to the user and confirm with him on those values. For instance, in the previous turns, the user was asking for an attraction in the centre and he is asking for a hotel now. In this case, don't directly assume that the hotel should be in the centre. Confirm with him.\\
- When the user is querying hotels, you should confirm with the user on the type of hotels instead of directly taking "hotel" as the type input value for query\_hotels API.\\
\\
\# Example with explanation:\\
<An example with explanation>\\
\midrule 

\end{tabular}
\end{scriptsize}
\end{table*}

%% file: acl_2025_sections/prompts/alfworld.tex
\subsection{AlfWorld Prompts}

\begin{table*}[!h]
\scriptsize
\begin{minipage}{0.9\linewidth}
\caption{A \react{} Main prompt for ALFWorld.}
\label{prompt:alfworld_react_main}
\centering
\ttfamily
\begin{tabular}{p\linewidth}
\midrule
You are a helpful assistant assigned with the task of problem-solving. You are adept at thinking, and acting. Your actions within the environment are strictly limited to a specific set of commands, and you must adhere to these rules without exception. \\
\\
1. Action Types:\\
   - THINK: Internal reasoning \\
   - ACT: Perform actions \\
\\
2. Response Format:\\
   You must choose only ONE of the following formats for your response:\\
   \\
   Think: [Your thought process]\\
   \\
   OR\\
\\
   Act: [The action you're taking]\\
\\
3. Available Actions for ACT Type:
\\
    Put\\
    GoTo\\
    Take\\
    Open\\
    Toggle\\
    Close\\
    Clean\\
    Heat\\
    Cool\\
    Use\\
    Look\\
\\
4. Grammar Rules for ACT Type:\\
    When suggesting or describing actions, you must strictly use the following grammar rules with special characters like '/' when necessary:\\
\\
    put \{object\} in/on \{receptacle\}\\
    go to \{receptacle\}\\
    take \{object\} from {receptacle}\\
    open \{receptacle\}\\
    toggle \{object\_or\_receptacle\}\\
    close \{receptacle\}\\
    clean \{object\} with \{receptacle\}\\
    heat \{object\} with \{receptacle\}\\
    cool \{object\} with \{receptacle\}\\
    use \{receptacle\}\\
    look\\
\\
You must always follow these rules precisely. Do not suggest or imply any actions outside of this defined set. \\

\midrule 

\end{tabular}
\end{minipage}%
\end{table*}

\begin{table*}[!h]
\scriptsize
\begin{minipage}{0.9\linewidth}

\caption{A \react{} prompt with In-context example on the ALFWorld clean task.}
\label{prompt:alfworld_im}
\centering
\ttfamily
\begin{tabular}{p\linewidth}
\midrule
Interact with a household to solve a task. Here are two examples. \\
You are in the middle of a room. Looking quickly around you, you see a armchair 2, a armchair 1, a bed 1, a countertop 1, a diningtable 1, a drawer 2, a drawer 1, a dresser 1, a garbagecan 1, a laundryhamper 1, and a sidetable 1.\\
Your task is to: put two creditcard in dresser.\\
> \textbf{\textcolor{Cerulean}{think}:} To solve the task, I need to find and take the first creditcard, then put it in dresser, then find and take the second creditcard, then put it in dresser.
OK.\\
> \textbf{\textcolor{Cerulean}{think}:} First I need to find the first creditcard. A creditcard is more likely to appear in drawer (1-2), coutertop (1), sidetable (1), diningtable (1), armchair (1-2), bed (1). I can check one by one, starting with drawer 1.\\
OK.\\
> go to drawer 1\\
The drawer 1 is closed.\\
> open drawer 1\\
You open the drawer 1. The drawer 1 is open. In it, you see a book 1, a cd 1, and a pen 1.\\
> go to drawer 2\\
The drawer 2 is closed.\\
> open drawer 2\\
You open the drawer 2. The drawer 2 is open. In it, you see nothing.\\
> go to countertop 1\\
On the countertop 1, you see a cellphone 2, a creditcard 4, a creditcard 3, a creditcard 2, a mirror 1, a pencil 2, and a pencil 1.\\
> \textbf{\textcolor{Cerulean}{think}:} Now I find the first creditcard (2). Next, I need to take it. I can find the second creditcard (3) later in countertop 1.\\
OK.\\
> take creditcard 2 from countertop 1\\
You pick up the creditcard 2 from the countertop 1.\\
> \textbf{\textcolor{Cerulean}{think}:} Now I take the first creditcard (2). Next, I need to put it in/on dresser 1.\\
OK.\\
> go to dresser 1\\
On the dresser 1, you see a mug 1, and a television 1.\\
> put creditcard 2 in/on dresser 1\\
You put the creditcard 2 in/on the dresser 1.\\
> \textbf{\textcolor{Cerulean}{think}:} Now I put the first creditcard in dresser. Next, I need to find the second creditcard. I can directly go to countertop 1.\\
OK.\\
> go to countertop 1\\
On the countertop 1, you see a cellphone 2, a creditcard 4, a creditcard 3, a mirror 1, a pencil 2, and a pencil 1.\\
> \textbf{\textcolor{Cerulean}{think}:} Now I find the second creditcard (3). Next, I need to take it.\\
OK.\\
> take creditcard 3 from countertop 1\\
You pick up the creditcard 3 from the countertop 1.\\
> \textbf{\textcolor{Cerulean}{think}:} Now I take the second creditcard (3). Next, I need to put it in/on dresser 1.\\
OK.\\
> go to dresser 1\\
On the dresser 1, you see a creditcard 2, a mug 1, and a television 1.\\
> put creditcard 3 in/on dresser 1\\
You put the creditcard 3 in/on the dresser 1.\\
\\
\textbf{<similar 2nd example>} \\

\midrule

\end{tabular}
\end{minipage}%
\end{table*}

\begin{table*}[!h]
\scriptsize
\begin{minipage}{0.9\linewidth}

\caption{A \respact{} Main prompt for ALFWorld.}
\label{prompt:alfworld_respact_main}
\centering
\ttfamily
\begin{tabular}{p\linewidth}
\midrule
You are a helpful assistant assigned with the task of problem-solving. You are adept at thinking, speaking and acting. Your actions within the environment are strictly limited to a specific set of commands, and you must adhere to these rules without exception. \\
\\
1. Action Types:\\
   - THINK: Internal reasoning \\
   - SPEAK: For Communication with User\\
   - ACT: Perform actions \\
\\
2. Response Format:\\
   You must choose only ONE of the following formats for your response:\\
   \\
   Think: [Your thought process]\\
   \\
   OR\\
   \\
   Speak: [Any communication, if necessary]\\
   \\
   OR\\
\\
   Act: [The action you're taking]\\
\\
3. Available Actions for ACT Type:
\\
    Put\\
    GoTo\\
    Take\\
    Open\\
    Toggle\\
    Close\\
    Clean\\
    Heat\\
    Cool\\
    Use\\
    Look\\
\\
4. Grammar Rules for ACT Type:\\
    When suggesting or describing actions, you must strictly use the following grammar rules with special characters like '/' when necessary:\\
\\
    put \{object\} in/on \{receptacle\}\\
    go to \{receptacle\}\\
    take \{object\} from {receptacle}\\
    open \{receptacle\}\\
    toggle \{object\_or\_receptacle\}\\
    close \{receptacle\}\\
    clean \{object\} with \{receptacle\}\\
    heat \{object\} with \{receptacle\}\\
    cool \{object\} with \{receptacle\}\\
    use \{receptacle\}\\
    look\\
\\
You must always follow these rules precisely. Do not suggest or imply any actions outside of this defined set. \\

\midrule 

\end{tabular}
\end{minipage}%
\end{table*}

\begin{table*}[!h]
\scriptsize
\begin{minipage}{0.9\linewidth}

\caption{A \respact{} In-Context Example on the ALFWorld put two task. The thoughts (1) decompose the goal, (2) track subgoal completion, (3) determine the next subgoal, and (4) reason via commonsense where to find an object and what to do with it. Speak actions demonstrate how the agent seeks clarification or guidance from the user, while environment actions show how the agent interacts with the Alfworld environment.}
\label{prompt:alfworld_respact}
\centering
\ttfamily
\begin{tabular}{p\linewidth}
\midrule

You are in the middle of a room. Looking quickly around you, you see a armchair 2, a armchair 1, a bed 1, a countertop 1, a diningtable 1, a drawer 2, a drawer 1, a dresser 1, a garbagecan 1, a laundryhamper 1, and a sidetable 1.\\
Your task is to: \textbf{put two creditcard in dresser.}\\
> \textbf{think:} To solve the task, I need to find and take the first creditcard, then put it in dresser, then find and take the second creditcard, then put it in dresser.
OK.\\
> \textbf{\textcolor{Cerulean}{think}:}  First I need to find the first creditcard. A creditcard is more likely to appear in drawer (1-2), coutertop (1), sidetable (1), diningtable (1), armchair (1-2), bed (1). Let me ask where to look for the creditcards.\\
OK.\\
> \textbf{\textcolor{ForestGreen}{speak}:} I need to find the first creditcard. A creditcard is more likely to appear in drawer (1-2), coutertop (1), sidetable (1), diningtable (1), armchair (1-2), bed (1). Where do you suggest I should look for the creditcard first?\\
> \textbf{\textcolor{BurntOrange}{Human}}:I think there are some credit cards on the countertop. Can you check there?\\
> go to countertop 1\\
On the countertop 1, you see a cellphone 2, a creditcard 4, a creditcard 3, a creditcard 2, a mirror 1, a pencil 2, and a pencil 1.\\
> \textbf{\textcolor{Cerulean}{think}:} Now I found three creditcards. creditcard (4), creditcard (3), creditcard (2). Let me ask which two I should pick.\\
OK.\\
> \textbf{\textcolor{ForestGreen}{speak}:} I found three creditcards. creditcard (4), creditcard (3), creditcard (2). Which two should I put in the dresser?\\
> \textbf{\textcolor{BurntOrange}{Human}}:Just creditcard 2 and creditcard 3. Please put them in the dresser.\\
> \textbf{\textcolor{Cerulean}{think}:} Now I find the first creditcard (2). Next, I need to take it. I can find the second creditcard (3) later in countertop 1.\\
OK.\\
> take creditcard 2 from countertop 1\\
You pick up the creditcard 2 from the countertop 1.\\
> \textbf{\textcolor{Cerulean}{think}:} Now I take the first creditcard (2). Next, I need to put it in/on dresser 1.\\
OK.\\
> go to dresser 1\\
On the dresser 1, you see a mug 1, and a television 1.\\
> put creditcard 2 in/on dresser 1\\
You put the creditcard 2 in/on the dresser 1.\\
> \textbf{\textcolor{Cerulean}{think}:} Now I put the first creditcard in dresser. Next, I need to take the second creditcard. I can directly go to countertop 1.\\
OK.\\
> go to countertop 1\\
On the countertop 1, you see a cellphone 2, a creditcard 4, a creditcard 3, a mirror 1, a pencil 2, and a pencil 1.\\
> \textbf{\textcolor{Cerulean}{think}:} Now I find the second creditcard (3). Next, I need to take it.\\
OK.\\
> take creditcard 3 from countertop 1\\
You pick up the creditcard 3 from the countertop 1.\\
> \textbf{\textcolor{Cerulean}{think}:} Now I take the second creditcard (3). Next, I need to put it in/on dresser 1.\\
OK.\\
> go to dresser 1\\
On the dresser 1, you see a creditcard 2, a mug 1, and a television 1.\\
> put creditcard 3 in/on dresser 1\\
You put the creditcard 3 in/on the dresser 1.\\

\midrule \\

\end{tabular}
\end{minipage}%
\end{table*}

\begin{table*}[!h]
\scriptsize
\begin{minipage}{0.9\linewidth}
\caption{A \respact{} Main prompt with Dialog Act Schema for ALFWorld.}
\label{prompt:alfworld_respact_schema}
\centering
\ttfamily
\begin{tabular}{p\linewidth}
\midrule
You are a helpful assistant assigned with the task of problem-solving. You are adept at thinking, and acting. Your actions within the environment are strictly limited to a specific set of commands, and you must adhere to these rules without exception. \\
\\
1. Action Types:\\
    think: Internal reasoning\\
    speak: For Communication with the User, limited to specific dialog acts\\
    act: Perform actions\\
\\
2. Response Format:\\
    You must choose only ONE of the following formats for your response:\\
\\
    think: [Your thought process]\\
\\
    OR\\
\\
    speak: [DialogAct]: [Any communication, if necessary]\\
\\
    OR\\
\\
    act: [The action you're taking]\\
\\
3. Available Actions for ACT Type:\\
    Put\\
    GoTo\\
    Take\\
    Open\\
    Toggle\\
    Close\\
    Clean\\
    Heat\\
    Cool\\
    Use\\
    Look\\

4. Grammar Rules for ACT Type:\\
    When suggesting or describing actions, you must strictly use the following grammar rules with special characters like '/' where necessary:\\
\\
        put \{object\} in/on \{receptacle\}\\
        go to \{receptacle\}\\
        take \{object\} from {receptacle}\\
        open \{receptacle\}\\
        toggle \{object\_or\_receptacle\}\\
        close \{receptacle\}\\
        clean \{object\} with \{receptacle\}\\
        heat \{object\} with \{receptacle\}\\
        cool \{object\} with \{receptacle\}\\
        use \{receptacle\}\\
        look\\
        
5. Dialog Act Rules for SPEAK Type:\\
    When communicating with the user, you must limit your communication to one of the predefined dialog acts. Choose an appropriate act based on the situation and communicate accordingly. The possible dialog acts are shown below as <dialog act>: purpose and its example:\\
\\
    <ReqForInstruction>: Ask for the next step if uncertain about the current task.\\
    Example: What should I do now?\\
\\
    <RequestOtherInfo>: Ask for additional details about the task.\\
    Example: Which 2 books should I pick?\\
\\
    <InfoObjectLocAndOD>: Provide information about the location or state of an object.\\
    Example: The knife 1 is on the countertop 1.\\
\\
    <ReqForObjLocAndOD>: Ask for the location or state of an object.\\
    Example: I am looking for a mug. Where is the mug?\\
\\
    <InformationOther>: Provide other relevant information.\\
    Example: I saw the pillow on the armchair.\\
\\
    <AlternateQuestions>: Provide alternative options to the user.\\
    Example: Which of the two creditcards. creditcard 1 or creditcard 2?\\
\\
    <Affirm>: Give affirmative responses.\\
    Example: Yes. I will proceed with that.\\
\\
    <Deny>: Give negative responses.\\
    Example: No. I don't think so.\\
\\
    <OtherInterfaceComment>: Provide comments related to the interface.\\
    Example: I am at the drawer 1. It is closed Should I open it?\\
\\
    <NotifyFailure>: Notify the user of a failure in completing a task.\\
    Example: Not able to do it. Please help\\

\midrule \\

\end{tabular}
\end{minipage}%
\end{table*}

%% file: acl_2025_sections/prompts/user_sim.tex
\subsection{User Simulator Prompts}

\begin{table*}[!h]
\scriptsize
\begin{minipage}{0.9\linewidth}
\caption{Prompt used for Helpful User Collaborating with Agent in Alfworld.}
\label{prompt:alfworld_user}
\centering
\ttfamily
\begin{tabular}{p\linewidth}
\midrule
You are a helpful user whose task is to guide an agent operating in the environment. You have knowledge of the objects necessary to complete the tasks and their where abouts as well as each step necessary for the agent in the environment to be successful which is as follows: \\
\\
You MUST respond ONLY when the agent speaks to you and ONLY regarding the specific information sought. You MUST NOT tell the agent any additional steps or information than what it seeks from you. Use a natural tone while reponding. \\
\\
Example: \\
\\
Oracle Information: ['go to dresser 1', 'take newspaper 1 from dresser 1', 'go to coffeetable 1', 'use desklamp 1'] \\
\\
Agent Query: I think a newspaper is more likely to appear in coffeetable (1-2), drawer (1-8), dresser (1-2), garbagecan (1), armchair (1), ottoman (1), sofa (1). Where do you suggest I should look for it first? \\
\\
Your Response: Hmm let me think. Can you please check the dresser 1? \\
\\
Think carefully and Provide your response for the following: \\
\\
Oracle Information: \{oracle\_text\}\\
\\
Agent Query: \\
\{query\}\\
\\
Your Response: \\
\midrule 

\end{tabular}
\end{minipage}%
\end{table*}


\begin{table*}[!h]
\scriptsize
\begin{minipage}{0.9\linewidth}
\caption{Prompt used for Perturbed Helpful User Simulator used for Alfworld.}
\label{prompt:alfworld_user}
\centering
\ttfamily
\begin{tabular}{p\linewidth}
\midrule
You are a human whose task is to guide an AI agent operating in a household environment. You have knowledge of the objects necessary to complete the tasks and their where abouts as well as each step necessary for the agent in the environment to be successful which is as follows: \\
\\
You MUST respond ONLY when the agent speaks to you and ONLY regarding the specific information sought. You MUST NOT tell the agent any additional steps or information than what it seeks from you. Use a natural tone while responding. \\
\\
Example: \\
\\
Oracle Information: ['go to dresser', 'take newspaper from dresser', 'go to coffeetable', 'use desklamp'] \\
\\
Agent Query: I think a newspaper is more likely to appear in coffeetable (1-2), drawer (1-8), dresser (1-2), garbagecan (1), armchair (1), ottoman (1), sofa (1). Where do you suggest I should look for it first? \\
\\
You responses should be ambiguous and Do NOT provide exact number of object or location. \\
\\
Your Response: Hmm let me think. Can you please check the dresser? \\
\\
Think carefully and Provide your response for the following: \\
\\
Oracle Information: \{oracle\_text\}\\
\\
Agent Query: \\
\{query\}\\
\\
Your Response: \\
\midrule 

\end{tabular}
\end{minipage}%
\end{table*}

\begin{table*}[!h]
\scriptsize
\begin{minipage}{0.9\linewidth}
\caption{Prompt used for UnHelpful User Collaborating with Agent in Alfworld.}
\label{prompt:alfworld_user}
\centering
\ttfamily
\begin{tabular}{p\linewidth}
\midrule
You are a user whose is interacting with an agent operating in the environment. You do not have knowledge of the objects necessary to complete the tasks and their where abouts in the environment for the agent to be successful, which is as follows: \\ 
\\
You MUST respond ONLY when the agent speaks to you and ONLY regarding the specific information sought. You MUST NOT tell the agent any additional steps or information than what it seeks from you. Use a natural tone while responding. \\
\\
Example: \\
\\
Agent Query: I think a newspaper is more likely to appear in coffeetable (1-2), drawer (1-8), dresser (1-2), garbagecan (1), armchair (1), ottoman (1), sofa (1). Where do you suggest I should look for it first? \\
\\
Your Response: Hmm I am not sure maybe check the ottoman? \\
\\
Think carefully and Provide your response for the following:\\
\\
Oracle Information: \{oracle\_text\}\\
\\
Agent Query: \\
\{query\}\\
\\
Your Response: \\
\midrule 

\end{tabular}
\end{minipage}%
\end{table*}

\begin{table*}[!h]
\scriptsize
\begin{minipage}{0.8\linewidth}
\caption{Prompt used for Helpful User Simulator used for Multiwoz}
\label{prompt:multiwoz_user}
\centering
\ttfamily
\begin{tabular}{p\linewidth}
\midrule

You are a dialogue simulator where you act as a user to talk to an AI assistant to complete some tasks.\\
You should carefully read and understand the User Goals below, then talk with the AI Assistant and gradually express the intents in the goals. Your purpose is to let the user achieve the goals as much as possible.\\
Note that the AI Assistant is not perfect. It may make various mistakes, including ignoring the user's requests, executing the wrong instructions, forgetting early conversation content, etc. The user you play should talk to the AI Assistant as patiently as possible, remind him to correct when you find that the AI assistant made a mistake, and complete the task as much as possible.\\
When asking some information of a venue (restaurant, hotel, attraction) or a train, you should specify the name or train id you choose.\\
When the dialogue goals are completed, you will output "Exit." to indicate the end of the dialogue. The you don't need to try conditions other than the dialogue goals.\\
You have a clear goal in mind, so you do not need to ask the AI assistant that "Is there anything else I need to know?".\\
You do not need to talk too much with the AI assistant. If the task goals are completed, please end the conversation as soon as possible.\\
There is also a reference dialogue example to achieve the goals. The simulated user may learn from the language style and dialogue strategy. The final simulated dialogue style should be similar to the reference dialogue style.\\
\\
\# An example is like this:\\
\\
You are given the goal of a dialogue:\\
\verb|`|\verb|`|\verb|`|\\
You are looking for a place to stay. The hotel should be in the cheap price range and should be in the type of hotel.\\
The hotel should include free parking and should include free wifi
Once you find the hotel you want to book it for 6 people and 3 nights starting from tuesday.\\
If the booking fails how about 2 nights.\\
Make sure you get the reference number.\\
\verb|`|\verb|`|\verb|`|\\
\\
You play the role of [User] and respond to the [Assistant]:\\
\verb|`|\verb|`|\verb|`|\\
\lbrack User]\\
I am looking for a place to stay that has a cheap price range it should be in a type of hotel.\\
\lbrack System]\\
Okay, do you have a specific area you want to stay in?\\
\lbrack User]\\
No, I just need to make sure it's cheap. Oh, and I need parking.\\
\lbrack System]\\
I found 1 cheap hotel for you that includes parking. Do you like me to book it?\\
\lbrack User]\\
Yes, please. 6 people for 2 nights starting on tuesday.\\
\lbrack System]\\
Booking was successful. reference number is: 7gawk763. Anything else I can do for you?\\
\lbrack User]\\
Exit.\\
\verb|`|\verb|`|\verb|`|\\
\\
Note that you don't include "[User]" in your response.\\
\\
\# User Goals for This Dialogue\\
\\
<user\_goals>\\

\midrule 

\end{tabular}
\end{minipage}%
\end{table*}

\begin{table*}[!h]
\scriptsize
\begin{minipage}{0.9\linewidth}
\caption{Prompt used for User Simulator used for Webshop}
\label{tab:webshop_user}
\centering
\ttfamily
\begin{tabular}{p\linewidth}
\midrule
User Profile: \{\{user\_profile\}\} \\
Your role is crucial in guiding the agent to make the right decision.\\
Remember: Your goal is to ask the agent to purchase one of the products from the search results from Agent Current Observation. If None of the search results match then you MUST ask the agent to pick from one of the closest available options. Here is an example:\\
\\
Shopping Goal: i want a noise cancelling cosycost usb microphone, and price lower than 60.00 dollars \\
\\
Conversation History: \\
\\
Agent Current Observation: Observation: \\
{[}Back to Search{]} \\
Page 1 (Total results: 50) \\
{[}Next >{]} \\
{[}B09L86RDXS{]} \\
Comfortable Bluetooth Headset, UX-M97 Wireless Headset with Microphone, Wireless Cell Phone Headset with Noise Isolation Mic Charging Base Mute Function for Xiaomi Poco F3 GT with Charging Dock \\
\$41.95 \\
{[}B092W6WNH4{]} \\
GAOMU IPX6 Waterproof Bluetooth Earbuds, True Wireless Earbuds, 20H Cyclic Playtime Headphones with Charging Case and mic for Android, in-Ear Stereo Earphones Headset for Sport Black \\
\$11.99 \\
{[}B014C9KQLM{]} \\
ASC Audio BlueTooth A2DP + USB Flash Drive Car Stereo Adapter Interface Compatible for Honda w/Navigation- Some Vehicles only- Compatible Vehicles Listed Below \\
\$84.95 \\
\\
Agent: It seems there are no results for a "noise cancelling Cosycost USB microphone" under \$60. Would you like me to search for a different brand or type of microphone? \\
Your Response: Let's proceed with the first one then. \\
\\
Now respond to this:\\
\\
Shopping Goal: \{instruction\}\\
\\
Conversation History: \{conversation\_history()\}\\
\\
Agent Current Observation: Observation: \{agent\_obs\}\\
\\
Agent: \{agent\_message\}\\
Your Response: \\
\midrule
\end{tabular}
\end{minipage}
\end{table*}